\documentclass[11pt]{article}

\usepackage[preprint]{acl}

\usepackage{times}
\usepackage{latexsym}
\usepackage[T1]{fontenc}
\usepackage[utf8]{inputenc}
\usepackage[protrusion=true,expansion=alltext,stretch=35,shrink=35]{microtype}
\usepackage{inconsolata}
\usepackage{graphicx}

\usepackage{booktabs}
\usepackage{amsmath}
\usepackage{amssymb}
\usepackage{amsfonts}

\usepackage{fontawesome5}   

\usepackage{xcolor}
\usepackage{array}
\usepackage{multirow}   
\usepackage{pgf}
\usepackage{subcaption}
\usepackage{xspace}

\newcommand{\csmot}[1]{\includegraphics[width=\linewidth]{case-studies/understanding/#1.png}}
\newcommand{\csq}[1]{\textit{``#1''}}
\definecolor{matchteal}{RGB}{0,121,107}
\newcommand{\gmatch}[1]{\textcolor{matchteal}{#1}}

\newcommand{\csgen}[2]{%
  \includegraphics[width=\linewidth]{case-studies/generation/#1.png}\par
  \vspace{-3pt}{\scriptsize $d{=}#2$\,m}%
}
\newcommand{\gcue}[1]{\textcolor{matchteal}{#1}}

\newcommand{\sysname}{MUGEN\xspace}

\newif\ifsectionpagedraft
\sectionpagedraftfalse

\usepackage{amsthm}

\AtBeginDocument{%
  \setlength{\abovedisplayskip}{5pt plus 2pt minus 2pt}%
  \setlength{\belowdisplayskip}{5pt plus 2pt minus 2pt}%
  \setlength{\abovedisplayshortskip}{3pt plus 2pt}%
  \setlength{\belowdisplayshortskip}{3pt plus 2pt}%
}

\renewcommand{\arraystretch}{0.96}
\makeatletter
\def\section{\@startsection{section}{1}{\z@}{-1.6ex plus -0.4ex minus -.2ex}%
  {1.0ex plus 0.2ex minus .2ex}{\large\bfseries\raggedright}}
\def\subsection{\@startsection{subsection}{2}{\z@}{-1.4ex plus -0.4ex minus -.2ex}%
  {0.5ex plus .2ex}{\normalsize\bfseries\raggedright}}
\def\paragraph{\@startsection{paragraph}{4}{\z@}{1.1ex plus 0.4ex minus .2ex}%
  {-1em}{\normalsize\bfseries}}
\makeatother

\usepackage{pifont}

\newcommand{\codeurl}{https://jye16.github.io/mugen-page/}
\newcommand{\hfurl}{https://huggingface.co/zy22b/MUGEN}
\newcommand{\hficon}{\raisebox{-0.25ex}{\includegraphics[height=1.05em]{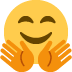}}}
\newcommand{\reslink}[3]{#1\,\href{#2}{\texttt{\small #3}}}

\title{\sysname: A Unified Framework for Efficient \underline{M}otion \underline{U}nderstanding and \underline{Gen}eration}

\author{
  \textbf{Zhankai Ye\textsuperscript{1}}\quad
  \textbf{Yukai Jin\textsuperscript{1}}\quad
  \textbf{Bingyang Wei\textsuperscript{2}}\quad
  \textbf{Bofan Li\textsuperscript{1}}\quad
  \textbf{Yusen Wu\textsuperscript{3}}\\
  \textbf{Fangyi Li\textsuperscript{4}}\quad
  \textbf{Shangqian Gao\textsuperscript{1}}\quad
  \textbf{Xin Liu\textsuperscript{1}}\\[2pt]
  {\normalfont\textsuperscript{1}Florida State University\quad
  \textsuperscript{2}Texas Christian University\quad
  \textsuperscript{3}University of Miami\quad
  \textsuperscript{4}University of Pennsylvania}\\[3pt]
  {\normalfont\reslink{\faGithub}{\codeurl}{\codeurl}\qquad
  \reslink{\hficon}{\hfurl}{\hfurl}}
}

\begin{document}

\maketitle

\begin{abstract}
  Grounding human motion in language, and language in motion, is a central step toward physical AI systems that can understand, generate, and communicate human behavior.
  Unified motion--language systems first coupled the two directions through a shared discrete motion codebook, but quantization limits generation quality.
  The strongest generators buy quality back at growing cost: stacked residual codebooks enlarge the representation; masked decoding stages, long autoregressive rollouts, and denoising chains of tens to hundreds of steps stretch inference; even the continuous-latent designs among them reach their latent only through an iterative diffusion head; and none of this decoding machinery serves understanding.
  We therefore propose \sysname, a unified motion--language framework that pays neither cost: \emph{no codebook, one draw}.
  A single adaptive-length autoencoder compresses any-length motion into a few continuous latent slots, the system's only motion representation: the language model generates them for text-to-motion and reads them back for motion understanding.
  Depth-routed hidden states let each slot read from the transformer depth it needs, and a calibrated head predicts a joint distribution over the full latent set, so a single draw carries the text-conditional, cross-slot variation a description permits.
  At a decoding cost of $K$ language-model steps, one draw, and one decoder pass, \sysname{} leads language-model baselines on FID on HumanML3D while raising retrieval precision above the real-motion reference under the standard evaluator, achieves the best CIDEr and BLEU@4 scores, and surpasses the discrete-token state of the art on every retrieval and alignment metric on SnapMoGen.
\end{abstract}


\section{Introduction}\label{sec:introduction}

Human motion provides a natural interface between language and embodied behavior. A general motion--language system should therefore support both directions of this interaction: generating plausible motions from textual descriptions and explaining observed motions in human language. Since these two tasks describe the same underlying behavior, it is desirable to handle them within a unified framework built around a shared motion representation. Existing systems such as TM2T and MotionGPT take an important step in this direction by representing motion through a VQ-VAE-style discrete interface and training a shared architecture for both text-to-motion generation and motion-to-text understanding~\cite{tm2t,jiang2023motiongpt}. However, the quantized codebook limits reconstruction and generation quality. More recent generators recover quality only at added cost: residual codebook hierarchies with staged masked prediction enlarge the motion representation, autoregressive motion-token decoding and diffusion-based denoising stretch inference to tens or hundreds of model passes, and even the continuous-latent designs among them must sample through an iterative diffusion head~\cite{momask,hwang2025snapmogen,t2mgpt,zhang2024motiondiffuse,zhu2025motiongpt3}. None of this added machinery serves the understanding branch. As a result, current systems can unify the language backbone, but high-quality and efficient motion generation and motion understanding still do not operate through the same motion interface.

This paper asks whether one continuous motion representation can support both generation and understanding without paying either cost.
Our insight is that motion does not need a discrete codebook to enter a language model.
A motion clip can instead become a small set of continuous latent slots that represents the sequence, gives the language model a compact generation target, and serves as the language model's input for motion understanding.

We present \sysname{}, a unified motion--language framework built around that interface under the opposite budget: \emph{no codebook, one draw}.
The budget itself is not new: early sequence-level VAEs such as TEMOS already decode a whole motion from one Gaussian draw~\cite{petrovich2022temos}, and the generators above outperform them at growing decoding cost.
What \sysname{} claims is the machinery that makes a single draw competitive again, inside a language model and on a representation that understanding shares.
Because the slots are continuous and learned jointly with the decoder rather than selected from a fixed vocabulary, the shared interface is no longer capped by codebook capacity, so unifying the two directions no longer sacrifices generation quality: joint training improves FID over a generation-only twin (Sec.~\ref{sec:kscaling}).
A single Adaptive-Length AutoEncoder (ALAE) maps any-length motion into $K$ continuous latent slots and decodes those slots back to frames.
For generation, the language model runs a $K$-step rollout seeded by a special \texttt{<MOT>} token, predicts the whole latent set, and sends one sampled latent set to the frozen decoder.
For understanding, the frozen encoder extracts the same $K$ slots from an observed motion, and a projector maps them into the language-model embedding space for caption generation.
Both directions share the language model, the motion interface, and the continuous representation.

\sysname{} makes this compact interface work with two design choices.
First, because the latent set is small, each slot must carry a large and distinct part of the sequence; if every slot reads the final layer, they all receive the same kind of evidence and can differ only by position.
Depth-routed hidden states remove this constraint by letting each slot read from the transformer depth it needs, with the final-layer interface retained as the special case where every slot routes to the last layer.
Second, a single draw must supply all the variation a description permits, and independent per-dimension noise cannot express variation shared across slots.
A calibrated low-rank factor head therefore predicts a joint distribution over the full latent set, so one draw carries text-conditional variation that spans slots.
Together, these components replace codebook stages, masked refinement, and denoising chains with $K$ language-model rollout steps, one draw, and one decoder pass.

We evaluate \sysname{} on HumanML3D and SnapMoGen under a fixed protocol.
The same sampler supplies all reported metrics, and test results follow the official protocols.
On HumanML3D, \sysname{} leads language-model baselines on FID, retrieval precision, and matching distance, and achieves the best CIDEr and BLEU@4 captioning scores, while masked-codebook pipelines keep the best FID.
On SnapMoGen, it further improves over the discrete-token state of the art on every retrieval rank and on CLIP alignment.

Our key contributions are as follows:

\noindent\hspace*{1em}$\bullet$\hspace{0.5em}
We show that high-quality motion generation and motion understanding can
share one compact continuous motion representation, without the codebook
hierarchies and iterative decoding that current generators pay for
quality.

\noindent\hspace*{1em}$\bullet$\hspace{0.5em}
\sysname{} realizes this interface with ALAE latent slots, depth-routed
hidden states that draw slot-specific evidence from the full depth of the
language model, and a calibrated low-rank latent head whose single draw
carries text-conditional variation across slots.

\noindent\hspace*{1em}$\bullet$\hspace{0.5em}
\sysname{} leads strong motion--language baselines on every retrieval
rank, on generation FID and matching distance, and on BLEU@4 and CIDEr,
while decoding each motion in $K$ language-model steps and one draw: $9$\,ms per motion, an
order of magnitude less inference compute, and $6$--$14\times$ lower
latency than the strongest masked-codebook and unified baselines in a
head-to-head measurement.


\section{Related Work}

\paragraph{Motion Generation.}
Text-to-motion generation is measured on paired benchmarks such as
HumanML3D~\cite{humanml3d} and the expressive whole-body
Motion-X~\cite{lin2023motionx}. Early continuous-latent models learn a
conditional motion distribution or align a motion autoencoder with CLIP
space~\cite{tevet2022motionclip,radford2021clip}; TEMOS, for instance,
decodes a whole motion from a single draw of a sequence-level
VAE~\cite{petrovich2022temos}.
Token-based systems cast motion as a discrete sequence: TM2T models the two
modalities reciprocally~\cite{tm2t}, T2M-GPT predicts motion tokens
autoregressively~\cite{t2mgpt}, AttT2M adds body-part and global--local
text attention over motion tokens~\cite{zhong2023attt2m}, and MotionGPT
shares a language-model
vocabulary~\cite{jiang2023motiongpt}, at a cost that scales with token count;
recent work scales tokenizer, model, and data
together~\cite{lu2025scamo,fan2025gotozero}. Diffusion models such as MDM
and MotionDiffuse instead denoise continuous
trajectories~\cite{ho2020ddpm,tevet2022human,zhang2024motiondiffuse},
optionally in a compressed latent space~\cite{rombach2022ldm}, with
retrieval-augmented~\cite{zhang2023remodiffuse},
physics-guided~\cite{yuan2023physdiff}, or
consistency-distilled~\cite{song2023consistency,dai2024motionlcm} variants.
Masked generative models refine tokens
iteratively~\cite{chang2022maskgit,pinyoanuntapong2024mmm,pinyoanuntapong2024bamm};
MARDM transfers masked autoregression to continuous latents sampled
through a diffusion head~\cite{mardm}.

These generators rest on a motion representation. Discrete designs follow
vector quantization~\cite{vqvae}, from single vocabularies to the
residual codebooks and staged decoding of MoMask~\cite{momask,lee2022rqvae} and
MoMask++~\cite{hwang2025snapmogen}. Our sampling budget of one draw and one
decoder pass predates all of these: TEMOS already paid it, though at a
quality every generator above has since surpassed. The closest
generation-only continuous design is MLD, which likewise compresses a clip
into a few latent tokens~\cite{chen2023mld}; that latent exists to host a diffusion process, so
it is reachable only by iterative denoising and never leaves the generator.
\sysname{} lets the language model write and read the slots directly, and
replaces the denoising chain with one draw from a predicted joint
distribution~\cite{sohn2015cvae}.  
Its claim is not the budget but what makes it competitive: behind $K$ cached language-model steps, 
depth-routed readout and a calibrated low-rank covariance bring single-draw continuous generation into the range of multi-stage discrete pipelines---ahead on retrieval and alignment, behind the masked-codebook family on FID---using the same slots read by the understanding branch.

\paragraph{Motion Understanding.}
TM2T first coupled captioning and generation in one tokenized
framework; MotionGPT trained a shared language model across
motion--language tasks, and MotionGPT3 replaced
discrete motion symbols with continuous features~\cite{zhu2025motiongpt3}.
UniMo couples the two directions through chain-of-thought supervision and
reinforcement post-training~\cite{wang2026unimo}. TMR
learns a contrastive text--motion space for retrieval~\cite{petrovich2023tmr},
and MG-MotionLLM addresses comprehension and generation at multiple
granularities~\cite{wu2025mgmotionllm}. GeoMotionGPT aligns the motion codebook
with the language-model embedding space through orthogonal sparse
projection~\cite{ye2026geomotiongpt}. These reciprocal systems couple
the two directions either through a shared discrete token interface,
whose quantization limits generation quality, or, in MotionGPT3,
through continuous latents reachable only by an iterative diffusion
head. \sysname{} keeps the shared continuous interface but reaches it
in one calibrated draw: one autoencoder yields latent slots serving as
generation target, understanding input, and decoder input.


\section{Methodology}



\subsection{Adaptive-Length AutoEncoder Without a Codebook}
\label{sec:approach}

\begin{figure}[t]
    \centering
    \includegraphics[width=0.88\columnwidth]{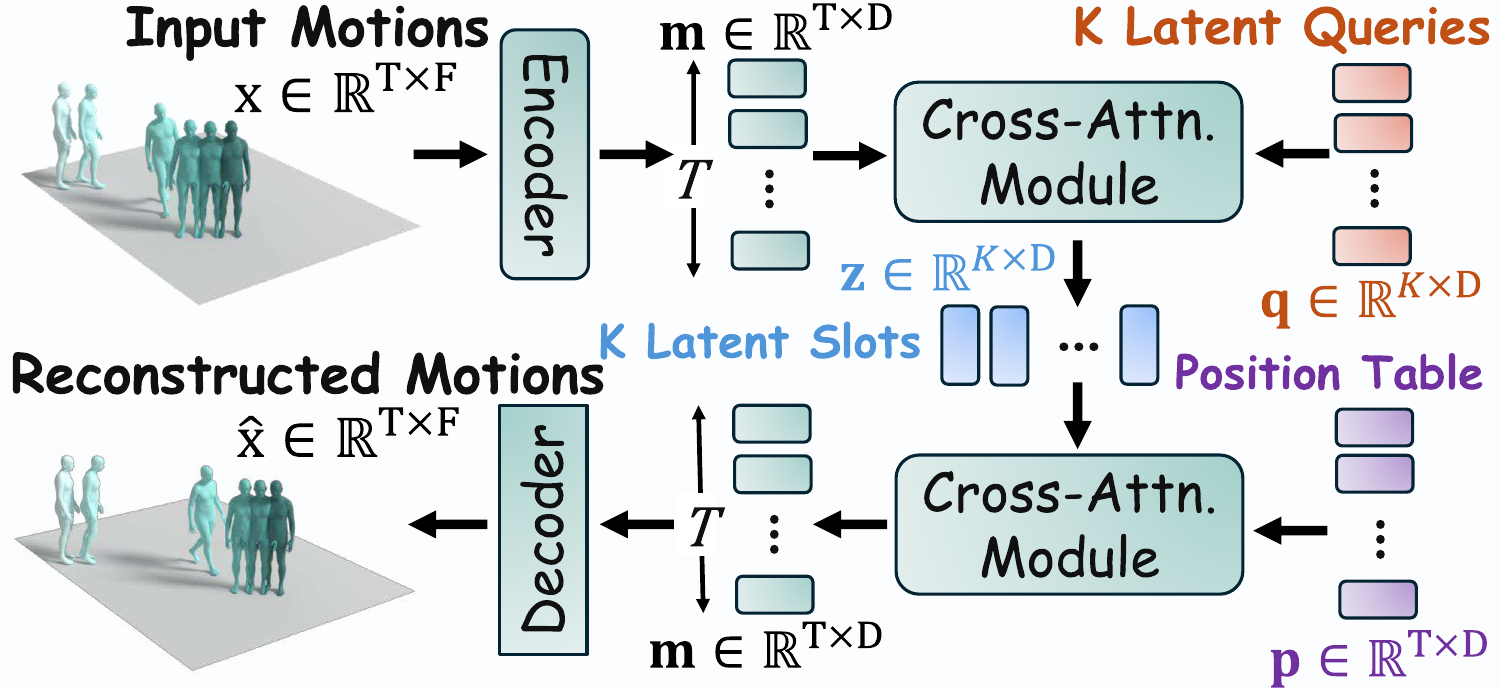}
    \vspace{-3mm}
    \caption{The Adaptive-Length AutoEncoder (ALAE).  The encoder compresses a
    motion of arbitrary length into exactly $K$ continuous latent slots by
    letting learned latent queries cross-attend to a temporal memory.  The decoder
    reconstructs the motion directly from those slots.}
    \vspace{-2mm}
    \label{fig:alae}
\end{figure}

Removing the codebook leads to two
requirements.  First, the representation must map variable-length motion into a
fixed budget.  Second, that budget must preserve enough structure for direct
reconstruction.  

We therefore build a continuous latent interface around a compact
autoencoding problem.  Given a normalized motion
sequence
$\mathbf{x}=[\mathbf{x}_1,\ldots,\mathbf{x}_T]\in\mathbb{R}^{T\times F}$,
where $F$ is the frame-feature dimension ($F{=}263$ for HumanML3D and
$F{=}296$ for SnapMoGen), the autoencoder reconstructs
$\hat{\mathbf{x}}\in\mathbb{R}^{T\times F}$ through exactly $K$ continuous
slots,
$\mathbf{z}=[\mathbf{z}_1,\ldots,\mathbf{z}_K]\in\mathbb{R}^{K\times D}$ with
$D{=}512$.  The latent budget $K$ is the only interface size exposed to later
modules.

The first step is to turn the input sequence into a temporal memory that keeps
local motion structure accessible.  ALAE (Fig.~\ref{fig:alae}) applies a
one-dimensional convolutional backbone with dilated ResNet blocks to produce
$\mathbf{m}_{1:T} \in \mathbb{R}^{T \times D}$, which preserves framewise
temporal structure while normalizing the feature space for the latent
bottleneck.

The second step is to compress this variable-length memory into a fixed set of
continuous latent slots.  Instead of assigning frames to discrete entries, learned
latent queries perform compression directly, following the fixed
latent-bottleneck principle of cross-attention
architectures~\cite{jaegle2021perceiver}. The construction is also related to
attention-based set encoders~\cite{lee2019settransformer} and learned slot
representations~\cite{locatello2020slotattention}; in the image domain, 1D
tokenizers likewise compress inputs of arbitrary resolution into a small,
controllable token budget~\cite{vibetoken2026}.  Let
$\mathbf{Q}=[\mathbf{q}_1,\ldots,\mathbf{q}_K]\in\mathbb{R}^{K\times D}$ be
the trainable latent queries.  A stack of cross-attention blocks updates these
queries by attending to the temporal memory:
\[
  \mathbf{z} = E_{\phi}(\mathbf{x}) =
  \mathrm{LN}\bigl(\mathrm{CA}_{\phi}(\mathbf{Q}, \mathbf{m}_{1:T})\bigr).
\]
Each block contains self-attention among the latent queries, cross-attention to
the temporal memory, and a feed-forward network.  The resulting slots summarize
complementary aspects of the motion rather than fixed temporal windows.  To
stabilize this specialization, an orthogonality penalty discourages
off-diagonal cosine similarity between query vectors:
\[
  \mathcal{L}_{\mathrm{orth}} =
  \frac{1}{K(K-1)} \sum_{i\neq j}
  {\left(
    \frac{\mathbf{q}_i^\top \mathbf{q}_j}
         {\|\mathbf{q}_i\|_2 \|\mathbf{q}_j\|_2}
  \right)}^2 .
\]
For $K{=}1$ the pair sum is empty and we define
$\mathcal{L}_{\mathrm{orth}}{=}0$.

The third step is to decode directly from the same slots.  The decoder samples
$T$ sinusoidal phase queries from a shared position table, so frame $t$ is
conditioned on its relative phase in the sequence rather than a fixed absolute
index.  These queries attend to $\mathbf{z}_{1:K}$ through cross-attention
blocks and are refined by a temporal convolutional decoder:
\[
  \hat{\mathbf{x}} =
  D_{\psi}(\mathbf{z}, T) =
  R_{\psi}\bigl(\mathrm{CA}_{\psi}(\mathbf{p}_{1:T}, \mathbf{z}_{1:K})\bigr).
\]
Because $K$ is fixed while $T$ is supplied explicitly, the autoencoder can train
on variable-length clips without changing the interface size seen by later
modules.

Training encourages the slots to remain reconstructive rather than merely
descriptive.  We use masked Smooth-L1 reconstruction over all frame features, a
joint-position Smooth-L1 term, a perceptual loss from a frozen
dataset-specific motion encoder, and the orthogonality penalty
$\mathcal{L}_{\mathrm{orth}}$.

After training, this autoencoder becomes the frozen motion interface.  The
checkpoint stores per-dimension latent mean and standard-deviation statistics;
the generator standardizes all latent targets with these statistics while
keeping the encoder and decoder fixed.  From this point on, the $K$ continuous
slots are the representation used by both generation and understanding.


\subsection{Latent Slot Generation with Depth Routing}

\begin{figure*}[t]
    \centering
    \includegraphics[width=0.80\textwidth]{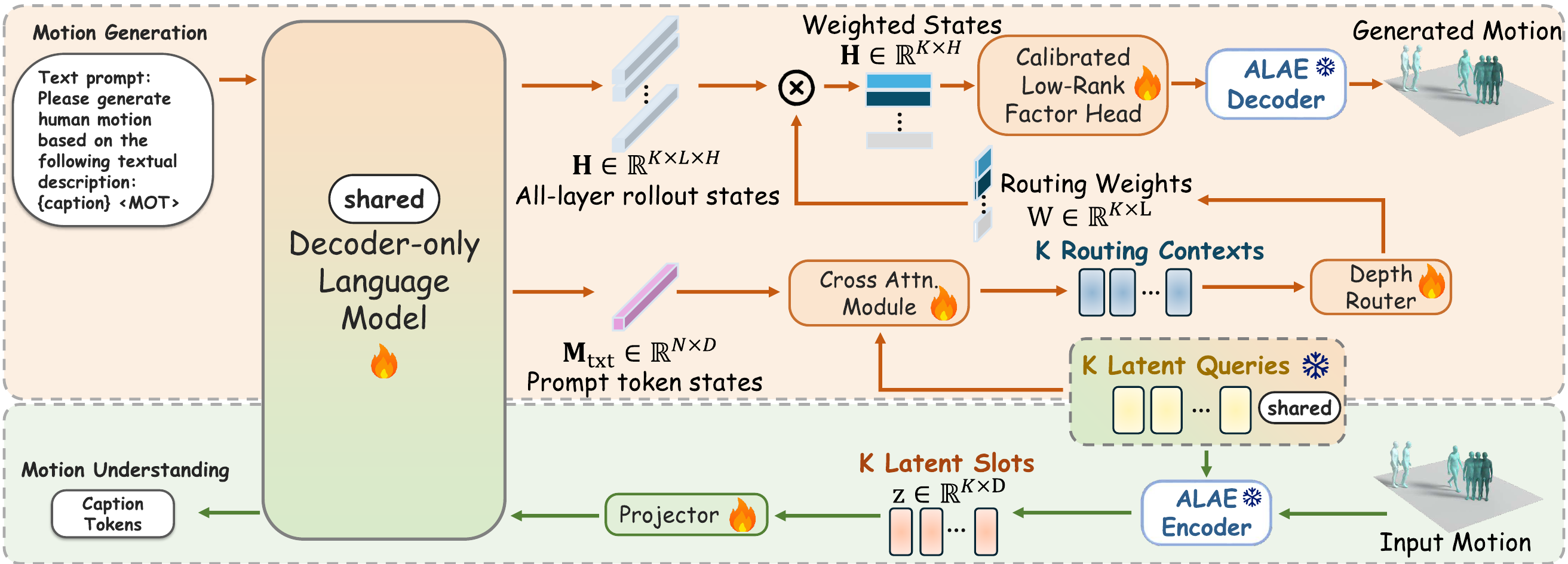}
    \caption{The shared motion--language interface of \sysname{}; flames mark
    trainable modules and snowflakes frozen ones.  \textbf{Generation (top):}
    a \texttt{<MOT>}-seeded $K$-step rollout keeps all $L$ layer states
    $\mathbf{H}$, a prompt-conditioned router assigns each latent slot its own
    depth mixture $\mathbf{W}$, and the calibrated low-rank factor head draws
    the complete latent set in a single shot for the frozen ALAE decoder.
    \textbf{Understanding (bottom):} the frozen ALAE encoder maps an input
    motion to the same $K$ slots, which a projector feeds to the same language
    model as caption context.}
    \label{fig:latent-generation}
    \vspace{-3mm}
\end{figure*}

Language-model motion generators read from the final transformer layer, an
interface inherited from next-token prediction rather than designed for
motion~\cite{t2mgpt,jiang2023motiongpt}, even though useful information is
distributed across depth~\cite{peters2018elmo,rogers2020bertology}.  The
mismatch sharpens in our setting: at the small budgets we target
(Sec.~\ref{sec:kscaling}), each of the $K$ slots carries a large fraction of
the sequence, and slots that all read the final layer are forced to draw on
the same kind of evidence, differing only by position.  We therefore let
each slot learn its own distribution over the $L$ layers; reading only the
final layer remains a special case, so depth routing strictly generalizes
the conventional interface.

\paragraph{Initializing continuous latent slots via autoregressive rollout.}\label{sec:rollout}
The generator must produce $K$ continuous latent slots that the frozen ALAE
decoder can turn into frames.  We therefore use the language model to create
these slots (Fig.~\ref{fig:latent-generation}).

Given a text description $y$, the prompt ends with a special \texttt{<MOT>} token whose
hidden state seeds a $K$-step rollout.  At step $k{=}1$ the model takes this
prompt state; at later steps it feeds the previous final-layer hidden state,
after a LayerNorm feedback transform, back to the language model as a
continuous input embedding. At each step, \sysname{} retains the output of
every transformer layer, producing a depth-indexed state tensor
$\mathbf{H} \in \mathbb{R}^{K \times L \times H}$, where $L$ is the number of
transformer layers and $H$ is the hidden width.  These states form the
evidence for latent prediction.  Which layers each latent slot reads
from is decided by the router described next.

\paragraph{Routing latent slots over depth.}\label{sec:whs}
Routing is conditioned on the prompt.  Latent queries imported from the ALAE
encoder cross-attend to the projected prompt states and yield one context vector
$\mathbf{c}_k\in\mathbb{R}^{D}$ per slot.  The router logits are then formed as
a \emph{bounded residual} rather than a direct regression,
\[
  \mathbf{e}_k =
  s_{\mathrm{s}}\tanh\!\bigl(\mathbf{g}_k / s_{\mathrm{s}}\bigr)
  \;+\;
  s_{\mathrm{d}}\tanh\!\bigl(\boldsymbol{\Delta}_k(y) / s_{\mathrm{d}}\bigr),
\]
where the free table $\mathbf{g}\in\mathbb{R}^{K\times L}$ carries the
prompt-independent part of the decision and
$\boldsymbol{\Delta}_k(y)=\mathrm{MLP}(\mathbf{c}_k)\in\mathbb{R}^{L}$ is a
per-prompt correction.  Squashing the two branches separately bounds the total
logit margin by $s_{\mathrm{s}}+s_{\mathrm{d}}$.  This matters because an
unconstrained router has a cheap way to suppress the sampling noise injected
during training: grow the margins until the softmax saturates.  Its Jacobian
then vanishes, the prompt branch stops receiving gradient, and routing freezes on
a prompt-independent pick it cannot escape.  The split also removes any reason to
spend the prompt branch on a constant, since $\mathbf{g}$ already represents one at
no cost to adaptivity.  We zero-initialize the correction head, so training
starts at the static solution and the static router acts as a performance floor
rather than a competing optimum.

Routing weights are Gumbel-Softmax samples~\cite{jang2017gumbel} during training
and a deterministic tempered softmax
$\mathbf{w}_k=\mathrm{softmax}(\mathbf{e}_k/\tau_{\mathrm{r}})$ at evaluation,
which keeps all reported metrics reproducible.  The state handed to the head is
the routed combination
\[
  \tilde{\mathbf{h}}_k \;=\; \sum_{l=1}^{L} w_{k,l}\, \mathbf{H}_{k,l},
  \qquad k = 1,\ldots,K .
\]
A bounded router is still free to ignore the prompt, so we add a small
mutual-information term that rewards routing decisively within a prompt and
differently across prompts.  Because the logits depend only on the prompt, they
are computed once per generation.

\paragraph{Producing latent slots with a calibrated low-rank factor head.}
The current architecture gives the model a single draw, so that draw must capture
the variation a text description permits.  A deterministic head therefore produces only
one motion per text description, while independent per-dimension noise cannot express
structured variation shared across latent slots.  We therefore predict a
structured latent distribution, implemented with a low-rank factor head over
the routed slot states.

The factor head reads the routed slot states: each
routed slot state $\tilde{\mathbf{h}}_k$ is layer-normalized and linearly
projected to the per-slot mean and diagonal scale, and the slot-pooled
normalized states produce the factor amplitudes.  It
models the flattened latent $\bar{\mathbf{z}} \in \mathbb{R}^{KD}$ with a
low-rank-plus-diagonal Gaussian
\[
  \begin{split}
  q\bigl(\bar{\mathbf{z}} \mid y\bigr) =
  \mathcal{N}\Bigl(
    \boldsymbol{\mu}(y),\;
    &\mathbf{U}\,\mathrm{diag}\bigl(\mathbf{a}(y)\bigr)^{2}\mathbf{U}^{\top}\\
    &+ \mathrm{diag}\bigl(\boldsymbol{\sigma}(y)^{2}\bigr)
  \Bigr),
  \end{split}
\]
where $\mathbf{U}\in\mathbb{R}^{KD\times r}$ is a learned global factor basis and
$\mathbf{a}(y)=\exp(\boldsymbol{\ell}(y))\in\mathbb{R}^{r}$ is a
text-conditioned amplitude vector, whose log-amplitudes are produced from the
slot-pooled hidden states by a linear layer. Because the basis spans \emph{across} slots, a single draw carries both within-slot and
cross-slot correlation.  As $\mathbf{a}\rightarrow 0$ the family degenerates to
the diagonal Gaussian, so the factor head strictly generalizes the diagonal
model.  Fitting its likelihood without stalling the mean requires the decoupled
anchor of Sec.~\ref{sec:latent-anchor}.

At inference a latent is drawn once, at a temperature $\tau$ that scales the
zero-mean perturbation around $\boldsymbol{\mu}$; the draw is mapped back
through the stored latent statistics, and the frozen ALAE decoder
turns it into $T$ frames.  Setting $\tau{=}0$ yields the deterministic
$\mu$-decode variant used as a reference protocol in our experiments.


\subsection{Joint Generation and Understanding}

The same continuous latent-slot interface serves both tasks.  Generation
predicts latent slots from text, while motion understanding reads the same slots
from the frozen encoder and maps them into the shared language model.  No
direction-specific tokenizer or decoder exists anywhere in the system.

\paragraph{Motion understanding branch.}
The frozen ALAE encoder converts an input motion into its $K$ latent slots, a
learned projector maps them into the language-model embedding space, and the
decoder-only language model is trained with teacher-forced caption cross-entropy loss. Label smoothing, a train-time Gaussian perturbation
of the input latents, and a flooding floor on the caption loss keep this branch
from overpowering the shared backbone late in training.  Both directions are
optimized on the same batches and share the language backbone, the
\texttt{<MOT>} interface, and the single motion representation.

\paragraph{Joint training objective.}\label{sec:latent-anchor}
During training, the motion branch decodes the predicted mean,
$\hat{\mathbf{x}}=D_{\psi}(\boldsymbol{\mu},T)$, and applies the same masked
reconstruction, joint-position, and perceptual terms used to train ALAE.  The full
loss adds the decoupled anchor, the motion-to-text (M2T) understanding
term, and the
router's mutual-information regularizer (Sec.~\ref{sec:whs}):
\[
  \begin{split}
  \mathcal{L} ={}&
  \mathcal{L}_{\mathrm{motion}}(\mathbf{x}, \hat{\mathbf{x}})
  + \lambda_{\mathrm{lat}}\,\mathcal{L}_{\mathrm{lat}}\\
  &+ \lambda_{\mathrm{m2t}}\,\mathcal{L}_{\mathrm{m2t}}
  + \lambda_{\mathrm{MI}}\,\mathcal{L}_{\mathrm{MI}} .
  \end{split}
\]
The decoupled anchor $\mathcal{L}_{\mathrm{lat}}$ is a Gaussian negative
log-likelihood on the target latents $\mathbf{z}^{\star}$, split into two terms
with complementary stop-gradients: one trains the mean under a detached diagonal
precision, the other fits the full low-rank covariance to the detached residual
by exact maximum likelihood.  Without that split, joint likelihood can explain
mean error as inflated variance, weakening the mean update and stalling the
motion prediction path.  Because the covariance parameters never enter the mean
term, the mean gradient stays stable while the amplitude $\mathbf{a}(y)$
self-calibrates without clamps or schedules.  No KL term is needed: likelihood
alone calibrates the predicted scale, and the learned
$\boldsymbol{\sigma}(y)$ becomes genuinely text-dependent.

The ALAE encoder and decoder stay frozen throughout, and the router, the factor
head, the feedback transform, the understanding projector, and the language
backbone are optimized jointly by this single objective.


\section{Experiments}\label{sec:experiments}

\subsection{Setup}\label{sec:setup}

\begin{table*}[t]
\centering
\small
\setlength{\tabcolsep}{2.4pt}
\begin{tabular*}{\textwidth}{@{\extracolsep{\fill}}llccccccc@{}}
\toprule
Type & Method & R@1$\uparrow$ & R@2$\uparrow$ & R@3$\uparrow$ & FID$\downarrow$ & MM-Dist$\downarrow$ & Diversity & MM \\
\midrule
& Real motions & 0.510 & 0.700 & 0.795 & -- & 3.012 & 9.503 & -- \\
\midrule
\multirow{7}{*}{Gen. only}
& MotionDiffuse & 0.491 & 0.681 & 0.782 & 0.630 & 3.113 & 9.410 & 1.55 \\
& T2M-GPT & 0.491 & 0.680 & 0.775 & 0.116 & 3.118 & 9.761 & 1.86 \\
& MMM & 0.504 & 0.696 & 0.794 & 0.080 & 2.998 & 9.411 & 1.16 \\
& MoMask & 0.521 & 0.713 & 0.807 & \textbf{0.045} & 2.958 & 9.620 & 1.24 \\
& BAMM & 0.525 & 0.720 & 0.814 & 0.055 & 2.919 & 9.717 & 1.69 \\
& MoMask++$^{\mathrm{in}}$ & 0.528 & 0.718 & 0.811 & 0.072 & 2.912 & -- & 1.23 \\
& MoMask++$^{\mathrm{cra}}$ & 0.517 & 0.709 & 0.803 & 0.069 & 2.948 & -- & 1.19 \\
\midrule
\multirow{4}{*}{Gen.\ \& Und.}
& MotionGPT & 0.492 & 0.681 & 0.778 & 0.232 & 3.096 & 9.528 & 2.01 \\
& MotionGPT3 & 0.553 & 0.747 & 0.837 & 0.208 & 2.725 & 9.700 & 1.02 \\
& UniMo & 0.539 & 0.738 & 0.831 & 0.177 & 2.768 & 10.042 & 1.92 \\
\cmidrule(l){2-9}
& \sysname~(Ours) & \textbf{0.579} & \textbf{0.772} & \textbf{0.859} & 0.087 & \textbf{2.629} & \textbf{9.510} & 0.83 \\
\bottomrule
\end{tabular*}
\vspace{-2mm}
\caption{Text-to-motion generation on HumanML3D. \sysname{} has the best
retrieval precision and matching distance in the table, both better than the
real-motion reference, the best FID among language-model generators, and the
Diversity closest to the real-motion reference ($9.510$ vs.\ $9.503$). The
masked-codebook pipelines keep the FID lead ($0.045$--$0.080$); the best of
their R@1 values trails \sysname{} by $0.051$.  Baseline numbers are
quoted from their original papers; MoMask++~\cite{hwang2025snapmogen}
appears as its in-context ($^{\mathrm{in}}$) and cross-attention
($^{\mathrm{cra}}$) conditioning variants.}
\label{tab:hml-main}
\end{table*}

\begin{table*}[t]
\centering
\small
\newcommand{\snapci}[2]{#1{\scriptsize$\pm#2$}}
\setlength{\tabcolsep}{1.7pt}
\resizebox{\textwidth}{!}{%
\begin{tabular}{llccccccc@{}}
\toprule
Type & Method & R@1$\uparrow$ & R@2$\uparrow$ & R@3$\uparrow$ & FID$\downarrow$ & CLIP score$\uparrow$ & Diversity & MM \\
\midrule
& Real motions & \snapci{0.940}{.001} & \snapci{0.976}{.001} & \snapci{0.985}{.001} & \snapci{0.001}{.000} & \snapci{0.837}{.000} & $19.5$--$19.8$ & -- \\
\midrule
\multirow{7}{*}{Gen. only}
& MDM & \snapci{0.503}{.002} & \snapci{0.653}{.002} & \snapci{0.727}{.002} & \snapci{57.783}{.092} & \snapci{0.481}{.001} & -- & \snapci{13.412}{.231} \\
& T2M-GPT & \snapci{0.618}{.002} & \snapci{0.773}{.002} & \snapci{0.812}{.002} & \snapci{32.629}{.087} & \snapci{0.573}{.001} & -- & \snapci{9.172}{.181} \\
& StableMoFusion~\cite{stablemofusion} & \snapci{0.679}{.002} & \snapci{0.823}{.002} & \snapci{0.888}{.002} & \snapci{27.801}{.063} & \snapci{0.605}{.001} & -- & \snapci{9.064}{.138} \\
& MARDM & \snapci{0.659}{.002} & \snapci{0.812}{.002} & \snapci{0.860}{.002} & \snapci{26.878}{.131} & \snapci{0.602}{.001} & -- & \snapci{9.812}{.287} \\
& MoMask & \snapci{0.777}{.002} & \snapci{0.888}{.002} & \snapci{0.927}{.002} & \snapci{17.404}{.051} & \snapci{0.664}{.001} & -- & \snapci{8.183}{.184} \\
& MoMask++$^{\mathrm{in}}$ & \snapci{0.805}{.002} & \snapci{0.904}{.002} & \snapci{0.938}{.001} & \snapci{15.56}{.071} & \snapci{0.684}{.001} & -- & \snapci{6.556}{.178} \\
& MoMask++$^{\mathrm{cra}}$ & \snapci{0.802}{.001} & \snapci{0.905}{.002} & \snapci{0.938}{.001} & \textbf{\snapci{15.06}{.065}} & \snapci{0.685}{.001} & -- & \snapci{7.259}{.180} \\
\midrule
Gen.\ \& Und.
& \sysname~(Ours)  & \textbf{\snapci{0.815}{.003}} & \textbf{\snapci{0.916}{.002}} & \textbf{\snapci{0.947}{.001}} & \snapci{21.05}{.074} & \textbf{\snapci{0.698}{.001}} & \snapci{19.79}{.037} & \snapci{6.09}{.320} \\
\bottomrule
\end{tabular}}
\vspace{-2mm}
\caption{Text-to-motion generation on SnapMoGen dataset. \sysname{} is the one entry
that also does understanding. It leads the
multi-stage discrete state of the art at every retrieval rank and on CLIP
score, and its Diversity is inside the ground-truth range. FID is the one
column where the residual codebook pipelines stay ahead.  Baseline numbers
are quoted from the SnapMoGen benchmark report; superscripts as in
Table~\ref{tab:hml-main}.}
\vspace{-2mm}
\label{tab:snap-main}
\end{table*}

We evaluate on two text-to-motion benchmarks that differ in scale, caption
style, and motion representation: \textbf{HumanML3D} (263-d frame features
at 20\,fps, short crowd-sourced captions) and \textbf{SnapMoGen} (longer,
more expressive captions), each with its own official splits and evaluator.
Generation is scored with FID~\cite{heusel2017fid}, R-Precision
(R@1/2/3)~\cite{rprecision}, matching distance (MM-Dist), Diversity, and
MultiModality (MM), plus SnapMoGen's CLIP score; understanding is scored
with BLEU~\cite{papineni2002bleu}, ROUGE, CIDEr~\cite{vedantam2015cider},
BERTScore~\cite{zhang2020bertscore}, and motion--text retrieval.  All test
numbers follow each benchmark's official 20-replication protocol; we report
means, with 95\% confidence intervals where space permits.

Stage one trains one ALAE per latent budget $K$ ($D{=}512$, four encoder and
four decoder layers).  Stage two freezes it and trains
GPT-2~\cite{gpt2} as the language backbone ($L{=}12$ layers) with the depth
router of Sec.~\ref{sec:whs} and the decoupled anchor of
Sec.~\ref{sec:latent-anchor} at factor rank $r{=}64$, learning generation
and understanding jointly on the same batches.  Sampling temperature and
checkpoints are selected on the validation split alone; the appendix gives
the full recipe and defines every metric.

The recipe is identical on both benchmarks and only the latent budget
differs ($K{=}2$ on HumanML3D, $K{=}4$ on SnapMoGen); on HumanML3D a single
checkpoint produces both the generation and the understanding numbers.
Swapping GPT-2 for Qwen3~\cite{qwen3} or SmolLM2~\cite{smollm2} with LoRA
adapters~\cite{lora} never surpasses the flagship's joint
generation--understanding score at up to $14\times$ the parameters, so we
keep the smallest backbone (ablation in the appendix).

\subsection{Motion Generation}\label{sec:eval-gen}

\begin{table*}[t]
\centering
\small
\setlength{\tabcolsep}{2.2pt}
\resizebox{\textwidth}{!}{%
\begin{tabular}{llccccccccc@{}}
\toprule
Type & Method & R@1$\uparrow$ & R@2$\uparrow$ & R@3$\uparrow$ & MM-Dist$\downarrow$ & BLEU@1$\uparrow$ & BLEU@4$\uparrow$ & ROUGE$\uparrow$ & CIDEr$\uparrow$ & BERTScore$\uparrow$ \\
\midrule
& Real motions & 0.523 & 0.725 & 0.828 & 2.901 & -- & -- & -- & -- & -- \\
\midrule
Und. only
& LaMP-M2T~\cite{li2025lamp} & 0.547 & -- & 0.831 & 2.808 & 47.8 & 13.04 & 37.1 & 28.9 & 32.7 \\
\midrule
\multirow{6}{*}{Gen.\ \& Und.}
& TM2T & 0.516 & -- & 0.823 & 2.935 & 48.9 & 7.00 & 38.1 & 16.8 & 32.2 \\
& MotionGPT & 0.543 & -- & 0.827 & 2.821 & 48.2 & 12.5 & 37.4 & 29.2 & 32.4 \\
& MoTe~\cite{wu2024mote} & 0.577 & -- & 0.871 & 2.649 & 46.7 & 11.15 & 37.4 & 31.5 & 30.3 \\
& MotionGPT3 & 0.573 & 0.773 & 0.864 & \textbf{2.426} & 59.083 & 19.412 & 46.173 & 28.721 & 35.231 \\
& UniMo & -- & -- & -- & -- & \textbf{63.10} & 19.74 & \textbf{48.8} & 46.69 & \textbf{54.26} \\
\cmidrule(l){2-11}
& \sysname~(Ours) & \textbf{0.586} & \textbf{0.784} & \textbf{0.873} & 2.536 & 60.753 & \textbf{21.980} & 47.367 & \textbf{50.406} & 43.369 \\
\bottomrule
\end{tabular}}
\caption{Motion understanding on the HumanML3D dataset. Rows are grouped by
what the reported model supports: the captioning-specific single-task model
LaMP-M2T against models that also generate motion. \sysname{} trains no
captioning-specific variant, yet leads every reported retrieval rank
together with BLEU@4 and CIDEr; UniMo, trained with curated
chain-of-thought annotations, leads BLEU@1, ROUGE, and BERTScore and
reports no retrieval columns.}
\label{tab:hml-m2t}
\end{table*}

\begin{table}[t]
\centering\small
\setlength{\tabcolsep}{2pt}
\begin{tabular*}{\columnwidth}{@{\extracolsep{\fill}}lrrrr@{}}
\toprule
Method & Params & GFLOPs & Latency & Throughput \\
 & (M) & /motion & (ms) & (motion/s) \\
\midrule
MoMask++ & 150.2 & 105.5 & 55 & 55.1 \\
MotionGPT3 & 299.7 & 95.5 & 136 & 53.2 \\
\sysname~(Ours) & 222.3 & 11.1 & 9 & 325.2 \\
\bottomrule
\end{tabular*}
\caption{Inference cost of generating one motion on HumanML3D.
\sysname{} spends $8.6$--$9.5\times$ less compute and responds
$6$--$14\times$ faster than MoMask++ and
MotionGPT3, with the highest batched throughput.}
\label{tab:compute}
\vspace{-2mm}
\end{table}

Tables~\ref{tab:hml-main} and~\ref{tab:snap-main} report generation, where
the strongest baseline on each is a multi-stage discrete codebook pipeline.
\sysname{} has the best text--motion correspondence in both tables, on
HumanML3D beyond the real-motion reference and on SnapMoGen ahead of both
MoMask++ variants at every rank with CIs separated; six of the ten
HumanML3D baselines also cross that reference by smaller margins, so part
of the effect belongs to the evaluator.  FID is the one metric where the
discrete pipelines stay ahead.  An oracle decode locates the cause in the
conditional sampler rather than the representation: adding half of the true
latent residual cuts SnapMoGen FID to $6.89$, while batch-shuffled
residuals collapse R@1 to $0.42$, so the missing variance is
text-conditional and directionally structured, and the gap is the price of
the single draw, not of removing the codebook.  Every column comes from one
calibrated sampler at the validation-selected temperature, which favors
fidelity; Diversity tracks the real-motion reference on both benchmarks, so
the low MultiModality reflects the operating point rather than a collapsed
distribution.

\subsection{Motion Understanding}\label{sec:eval-m2t}

\begin{table}[t]
\centering
\small
\setlength{\tabcolsep}{4pt}
\resizebox{\columnwidth}{!}{%
\begin{tabular}{lcccccc@{}}
\toprule
HumanML3D & R@1$\uparrow$ & R@2$\uparrow$ & R@3$\uparrow$ & FID$\downarrow$ & Div. & MM \\
\midrule
\multicolumn{7}{@{}l}{\emph{Latent budget $K$ (depth-routed rollout)}} \\
$K{=}1$ & 0.574 & 0.771 & 0.859 & 0.109 & 9.50 & 0.67 \\
$K{=}2$ & \textbf{0.579} & 0.772 & 0.859 & \textbf{0.087} & 9.51 & 0.83 \\
$K{=}4$ & 0.577 & \textbf{0.774} & \textbf{0.862} & 0.091 & 9.46 & 0.70 \\
$K{=}8$ & 0.567 & 0.763 & 0.853 & 0.111 & 9.45 & 0.91 \\
$K{=}16$ & 0.566 & 0.761 & 0.852 & 0.143 & 9.46 & 0.74 \\
\midrule
\multicolumn{7}{@{}l}{\emph{Rollout interface, at the $K{=}2$ budget}} \\
Last layer only & 0.574 & 0.770 & 0.860 & 0.123 & -- & 1.05 \\
\bottomrule
\end{tabular}}
\caption{Latent budget and rollout depth on the HumanML3D test split.
Retrieval is flat over $K{=}1$--$4$ and falls beyond it, while FID is best
at the small budgets.}
\vspace{-4mm}
\label{tab:kscaling}
\end{table}

Table~\ref{tab:hml-m2t} evaluates the understanding branch, which reads the
same frozen latent slots the generation branch predicts.  \sysname{} leads
every reported retrieval rank, above the real-motion reference, together
with BLEU@4 and CIDEr, at no captioning-specific training cost.  UniMo,
trained on curated chain-of-thought annotations, leads the remaining
caption metrics but reports no motion-side retrieval, and matching distance
is the one column where a baseline stays ahead.  SnapMoGen understanding
has no published baseline and is reported in the appendix.

\vspace{-2mm}
\subsection{Inference Cost}\label{sec:eval-cost}
\vspace{-2mm}
\sysname{} reaches the quality above with $K$ cached language-model steps,
one structured Gaussian draw, and one decoder pass, a budget no baseline in
Tables~\ref{tab:hml-main} and~\ref{tab:snap-main} matches.
Table~\ref{tab:compute} quantifies the gap on a single NVIDIA L4, with each
public checkpoint at the operating point of its reported results.  The
advantage is structural: MoMask++ runs $18$ masked-decoding steps across
four residual scales with two guided passes per step, MotionGPT3 emits its
motion latents one position at a time through an iterative diffusion head,
and diffusion pipelines take tens to hundreds of denoising steps.  Batching
narrows the latency gap to about $6\times$ without closing it.  The saving
lies in inference steps rather than model size: \sysname's parameter count
and peak memory both sit between the two baselines'.  The measurement
protocol and comparison scope are in the appendix.

\subsection{Ablations}\label{sec:kscaling}

Table~\ref{tab:kscaling} varies the two structural choices of the rollout:
the latent budget $K$, which sets the ``codebook'' size and, with key-value
caching, the inference step count; and the depth at which the slots are
read out.

\paragraph{A small budget suffices, on both benchmarks.}
Retrieval holds over $K{=}1$--$4$ and falls beyond it on HumanML3D;
SnapMoGen agrees, with every retrieval metric and the CLIP score peaking at
$K{=}4$ (table in the appendix).  FID separates the two: best at the small
budgets on HumanML3D, best at $K{=}16$ on SnapMoGen over a far narrower
spread, a gain that costs $1.4$ R@1 points.  Since inference cost grows
with $K$, we take the smallest budget that holds quality,
$K{=}2^{\dagger}$ on HumanML3D and $K{=}4$ on SnapMoGen.
Calibrated sampling beats $\mu$-decode on FID at every budget on both
datasets, at a cost of ${\approx}0.6$ R@1 points, and $\tau{=}1.0$ is never
validation-optimal (the appendix reports the full sweep).

\paragraph{Routing the read-out over depth.}
Depth routing improves FID by $29\%$ at fixed budget
($0.123\rightarrow0.087$), more than any move inside the $K{=}1$--$4$
region above, and the two slots take complementary roles
(Fig.~\ref{fig:router-slots}): $k_0$ keeps $98\%$ of its routing mass in
layers~9--11, while $k_1$ scans layers~5--11.  They pick the same top layer
on only $49\%$ of test prompts, so the pair is not a redundant clone.  The
router adds a small cross-attention stack and MLP over per-layer states the
backbone has already computed.

\begin{table}[tb]
\centering
\footnotesize
\newcommand{\uci}[2]{#1{\scriptsize$\pm#2$}}
\setlength{\tabcolsep}{2.0pt}
\resizebox{\columnwidth}{!}{%
\begin{tabular}{@{}lccccc@{}}
\toprule
Training & FID$\downarrow$ & R@1$\uparrow$ & CIDEr$\uparrow$ & BERT$\uparrow$ & Params \\
\midrule
Gen.\ only & \uci{0.107}{.004} & \uci{0.578}{.002} & -- & -- & 221.9M \\
Und.\ only & -- & -- & 52.8 & \textbf{44.8} & 222.3M \\
Joint, no floor & \uci{0.100}{.004} & \uci{0.574}{.002} & 40.7 & 39.2 & 222.3M \\
Joint, $\lambda{=}0.25$ & \uci{0.089}{.003} & \uci{0.576}{.002} & \textbf{53.5} & 44.7 & 222.3M \\
Joint, $\lambda{=}1$$^{\dagger}$ & \uci{\textbf{0.087}}{.004} & \uci{\textbf{0.579}}{.002} & 50.4 & 43.4 & 222.3M \\
Joint, $\lambda{=}4$ & \uci{0.111}{.003} & \uci{0.571}{.002} & 49.7 & 43.2 & 222.3M \\
\bottomrule
\end{tabular}}
\setlength{\abovecaptionskip}{4pt}
\caption{Unified-training ablation on the HumanML3D test set; $\lambda$ is
the captioning weight $\lambda_{\mathrm{m2t}}$, $^{\dagger}$ the reported
flagship, and BERT abbreviates BERTScore.}
\label{tab:unified}
\vspace{-2mm}
\end{table}

\paragraph{Unified training improves generation.}
Table~\ref{tab:unified} varies the training objective at the fixed
flagship recipe.  Dropping the captioning branch makes generation worse,
not better (FID $0.107$ vs.\ $0.087$, disjoint confidence intervals,
unchanged R@1): the understanding branch regularizes the shared motion
representation.  The reverse cost is small: the flagship stays within
$5\%$ of a dedicated captioner's CIDEr while adding only a $0.39$M
projector ($+0.18\%$), where task-specialized models would double the
footprint.  Removing the
caption-loss floor degrades both directions at once.


\section{Conclusion}\label{sec:conclusion}

We presented \sysname, a unified motion--language framework that replaces
discrete motion tokens with a small set of continuous latent slots shared
across generation and understanding. Depth-routed latent extraction and a
calibrated low-rank factor head make this interface effective at a decoding
cost of $K$ language-model steps, one draw, and one decoder pass, an order
of magnitude below the measured inference cost of the strongest multi-stage
baselines. \sysname{} delivers strong and
consistent results across retrieval, alignment, generation, and captioning
benchmarks. These results show that a unified system does not need to buy
retrieval, alignment, and captioning quality with codebook hierarchies or
iterative decoding---FID is the one remaining gap to the masked-codebook
family---and that a continuous latent interface is a simple and
competitive foundation for motion--language modeling.

\bibliography{citation}


\appendix

\section{Experimental Setup Details}
\label{app:setup}

This appendix expands the experimental setup of the main paper.

\paragraph{Implementation.}
Stage one trains one ALAE per latent budget $K$ (500 epochs, latent
dimension $D{=}512$, four encoder and decoder layers).  Stage two uses GPT-2
\cite{gpt2} as the language backbone ($L{=}12$ layers) and trains for 500
epochs with the decoupled anchor of the main paper; the
factor rank is $r{=}64$ for every $K$ and both datasets.  The depth router
uses two cross-attention blocks over the frozen
ALAE latent queries and a shared two-layer MLP; its Gumbel temperature anneals
from $5$ to $1.5$ over the first 100 epochs, and evaluation always uses the
deterministic tempered mixture.
Generation and understanding are trained
jointly on the same batches throughout.  The ALAE encoder and decoder are
frozen in stage two.  All configurations of the $K$-scaling study are
generated from a single base configuration with $K$ as the only change.

\paragraph{Computing infrastructure.}
All experiments ran on a SLURM-managed academic GPU cluster under Linux.
Each training run, for both stages and both datasets, used one NVIDIA B200
GPU, 16 CPU cores, and 160\,GB of system memory; no experiment uses more
than a single GPU.  A stage-one run completes in under 6 GPU-hours; a full
500-epoch stage-two run takes about 6 GPU-hours on HumanML3D and up to
about 21 GPU-hours on SnapMoGen at the largest latent budget.  Evaluation
jobs run in the same environment on a single B200 or L4 GPU.  The software
stack is Python 3.11, a PyTorch 2.11 nightly build with CUDA 12.8 (required
by the B200 architecture), PyTorch Lightning 2.0, and Hugging Face
Transformers 4.47 (4.51 for the backbone ablation of App.~\ref{app:backbone},
the minimum version that supports Qwen3); the released code pins the exact
version of every dependency.  The main paper compares the inference
cost of the reported system against the two baselines on one shared GPU, and
App.~\ref{app:compute} records the full measurement protocol.

\paragraph{Protocol and model selection.}
The sampling temperature is selected on the validation split only, and the
test split is evaluated at the selected operating point under the official
20-replication protocol (means with 95\% confidence intervals), with
$\tau{=}1.0$ and $\mu$-decode reported as references on HumanML3D
(Table~\ref{tab:hml-sampling}).  Model selection likewise uses only the
sampled validation split: the unified HumanML3D flagship takes the
checkpoint with the best joint generation--understanding score of
Eq.~\ref{eq:joint-score}, so a
single checkpoint serves both the generation and the understanding results
reported for HumanML3D in the main paper, while the $K$-scaling
study and the SnapMoGen flagship take the best-FID checkpoint.  The temperature-calibration
procedure itself is detailed in App.~\ref{sec:sampling-protocol}.

\paragraph{The joint selection score.}
Every metric is first mapped to a higher-is-better $0$--$100$ scale.
Generation contributes four terms: retrieval
$S_{\mathrm{R}}{=}100\,(\mathrm{R@1}{+}\mathrm{R@2}{+}\mathrm{R@3})/3$;
fidelity $S_{\mathrm{F}}{=}100\,\tau_f/(\tau_f{+}\mathrm{FID})$ with
$\tau_f{=}1$, a rational map with no saturation regime; matching
$S_{\mathrm{M}}{=}100\,\mathrm{MMDist}_{\mathrm{gt}}/\mathrm{MMDist}$, which
rises above $100$ when the model beats the real-motion reference (on
SnapMoGen the evaluator's matching score is a similarity, so the ratio is
inverted); and diversity
$S_{\mathrm{D}}{=}100\,\max\bigl(0,\,1{-}\lvert\mathrm{Div}{-}\mathrm{Div}_{\mathrm{gt}}\rvert/\mathrm{Div}_{\mathrm{gt}}\bigr)$
with $\mathrm{Div}_{\mathrm{gt}}{=}9.5$ on HumanML3D, which penalizes
deviation from the reference in either direction.  They aggregate as the
weighted geometric mean
$\mathrm{Avg}_G{=}\bigl(S_{\mathrm{R}}^{2}\,S_{\mathrm{F}}^{2}\,
S_{\mathrm{M}}\,S_{\mathrm{D}}\bigr)^{1/6}$, so no term can compensate for
another.  Understanding contributes the plain benchmark average
$\mathrm{Avg}_U$ of the motion-to-text retrieval mean
$100\,(\mathrm{R@1}{+}\mathrm{R@2}{+}\mathrm{R@3})/3$ and
$100{\times}$BLEU@1, BLEU@4, ROUGE, and CIDEr; BERTScore joins this mean
only where it is computed, and it is disabled during validation for cost,
so selection uses the five-term mean.  The selected checkpoint maximizes
\begin{equation}
  S=\mathrm{Avg}_G^{\,\alpha}\,\mathrm{Avg}_U^{\,1-\alpha},
  \qquad \alpha=0.5,
  \label{eq:joint-score}
\end{equation}
with every geometric-mean factor, in $\mathrm{Avg}_G$ and in
Eq.~\ref{eq:joint-score}, clamped below at $10^{-6}$: the geometric mean
ensures that degrading either direction degrades $S$, so the saved
checkpoint must serve both.

\section{Evaluation Metrics: Definitions and Motivation}
\label{app:metrics}

All metrics are computed in the feature space of each benchmark's official
evaluator: the contrastive text--motion evaluator of \citet{humanml3d} on
HumanML3D, and the TMR-style dual encoder of
\citet{hwang2025snapmogen} on SnapMoGen~\cite{petrovich2023tmr}.  Let
$f(\cdot)$ and $g(\cdot)$ denote the evaluator's motion and text encoders.
Using each benchmark's official evaluator keeps every number directly
comparable with the published baselines.

\paragraph{FID~\cite{heusel2017fid}.}
Gaussians $(\boldsymbol{\mu}_r,\boldsymbol{\Sigma}_r)$ and
$(\boldsymbol{\mu}_g,\boldsymbol{\Sigma}_g)$ are fitted to the evaluator
features of the real and generated motions, and
\[
  \mathrm{FID} =
  \lVert \boldsymbol{\mu}_r - \boldsymbol{\mu}_g \rVert_2^2
  + \operatorname{Tr}\!\bigl(\boldsymbol{\Sigma}_r + \boldsymbol{\Sigma}_g
  - 2\,(\boldsymbol{\Sigma}_r \boldsymbol{\Sigma}_g)^{1/2}\bigr).
\]
FID measures distributional fidelity: whether the generated set as a whole
occupies the same region of feature space as real motion.

\paragraph{R-Precision~\cite{rprecision}.}
Each generated motion is scored against a pool containing its ground-truth
description and randomly mismatched ones (pool size 32 on HumanML3D and 100
on SnapMoGen, following each benchmark), ranked by the Euclidean distance
between $f(\hat{\mathbf{x}})$ and $g(y)$.  R@$k$ is the fraction of samples
whose true description ranks in the top $k$; it measures per-sample
text--motion correspondence.

\paragraph{Matching distance (MM-Dist).}
The mean Euclidean distance $\lVert f(\hat{\mathbf{x}}) - g(y) \rVert_2$
between each generated motion and its own description, a threshold-free
companion to R-Precision.

\paragraph{CLIP score (SnapMoGen).}
The cosine similarity between the $\ell_2$-normalized motion and text
embeddings of the benchmark's CLIP-style dual
encoder~\cite{radford2021clip}, averaged over the test set.

\paragraph{Diversity and MultiModality.}
Diversity is the mean pairwise feature distance over 300 random pairs of
generated motions; it should match the real-motion reference, since values
far below or above it indicate a mismatch with the true output spread.
MultiModality is the mean pairwise distance among 30 motions generated for
the same description, averaged over 100 descriptions; it measures the
within-text variation that remains after conditioning.

\paragraph{Captioning metrics.}
BLEU@$n$~\cite{papineni2002bleu} is modified $n$-gram precision with a
brevity penalty; ROUGE is a recall-oriented measure based on the longest
common subsequence; CIDEr~\cite{vedantam2015cider} is a TF-IDF-weighted
$n$-gram consensus score; BERTScore~\cite{zhang2020bertscore} matches
contextual embeddings and captures semantic similarity beyond exact
$n$-gram overlap.  The retrieval columns of the understanding table use the
evaluator above with generated captions in place of the ground truth.

\paragraph{Why this set.}
Together these metrics separate the three requirements a text-to-motion
system must satisfy at once: distributional fidelity (FID), per-sample text
correspondence (R-Precision, MM-Dist, CLIP score), and variation (Diversity,
MultiModality).  The captioning side combines lexical evidence (BLEU, ROUGE,
CIDEr) with semantic evidence (BERTScore).  This is also the standard
reporting set of both benchmarks, which keeps every comparison in the
main paper on published ground.

\section{Complete Hyperparameters}
\label{app:hparams}

\begin{table}[t]
\centering
\small
\setlength{\tabcolsep}{3pt}
\resizebox{\columnwidth}{!}{%
\begin{tabular}{@{}ll@{}}
\toprule
Hyperparameter & Value \\
\midrule
\multicolumn{2}{@{}l}{\emph{Stage one: ALAE}} \\
Optimizer & AdamW \\
Weight decay & $10^{-4}$ (none on bias/norm) \\
Learning rate; schedule & $3{\times}10^{-4}$; cosine, 3\% warmup \\
Batch size; epochs & 512; 500 \\
Slot dim $D$; enc./dec.\ layers & 512; 4 / 4 \\
Heads; FFN; dropout & 8; 2048; 0.05 \\
Conv backbone & depth 3, 2 res blocks, dilation 3 \\
$\lambda_{\mathrm{rec}}/\lambda_{\mathrm{joint}}/\lambda_{\mathrm{perc}}/\lambda_{\mathrm{orth}}$ & $1/0.5/10/1$ \\
\midrule
\multicolumn{2}{@{}l}{\emph{Stage two: language model}} \\
Backbone & GPT-2 base ($L{=}12$) \\
Frozen modules & ALAE enc./dec.; text embeddings \\
Optimizer & AdamW, $\beta{=}(0.9,0.99)$, decay 0.01 \\
Learning rate; schedule & $10^{-4}$; 5-ep.\ warmup, cosine \\
Batch; epochs; precision & 256\,/\,$128{\times}2$; 500; fp32 \\
Max sequence length & 256 \\
Router & 2 cross-attn blocks, MLP 512 \\
Router $s_{\mathrm{s}}$, $s_{\mathrm{d}}$; $\lambda_{\mathrm{MI}}$ & 4, 4; 0.05 \\
Gumbel temperature & $5 \rightarrow 1.5$, first 100 epochs \\
Factor rank $r$; anchor $\lambda_{\mathrm{lat}}$ & 64; 2 \\
$\lambda_{\mathrm{rec}}/\lambda_{\mathrm{joint}}/\lambda_{\mathrm{perc}}$ & $1/0.5/10$ \\
$\lambda_{\mathrm{m2t}}$; label smoothing & 1; 0.1 \\
M2T latent noise; loss floor & 0.05; 2.8 \\
M2T decoding & greedy, $\leq 64$\,/\,$128$ tokens \\
\midrule
\multicolumn{2}{@{}l}{\emph{Evaluation protocol}} \\
$\tau$ sweep (validation only) & $\tau\in\{0.5,0.55,\ldots,1.0\}$ \\
Test protocol & 20 replications, batch 128 \\
Random seed & 1234 (all runs) \\
\bottomrule
\end{tabular}}
\caption{Complete final hyperparameters. Slash-separated values are
HumanML3D\,/\,SnapMoGen; every other value is shared across both datasets and
all latent budgets $K$.}
\label{tab:hparams}
\end{table}

Table~\ref{tab:hparams} lists every final hyperparameter of the recipe that
produces the reported results.  SnapMoGen reaches the same effective batch
size as HumanML3D through two-step gradient accumulation over batches of
128.  All runs fix the global random seed to 1234
through Lightning's \texttt{seed\_everything}, and the released
configuration files reproduce every row unchanged.

\section{Router Implementation Details}
\label{app:router}

This appendix expands the depth-routed weighted hidden states of the main paper.

\paragraph{Prompt memory.}
Let $\mathbf{h}^{(L)}_{1:N} \in \mathbb{R}^{N \times H}$ be the final-layer
states over the $N$ prompt positions.  We project the entire sequence to the
latent width and retain it as memory,
$\mathbf{M}_{\mathrm{txt}} = \mathbf{W}_{\mathrm{mem}}\,\mathbf{h}^{(L)}_{1:N}
\in \mathbb{R}^{N \times D}$, and supply the padding mask to the attention so
that only real tokens are attended.  Pooling the prompt into a single vector
instead would leave the cross-attention with one key.  Its softmax is then
identically one, the block degenerates into a gated linear map of that vector,
and the queries lose the ability to look up \emph{which} words they should route
on.

\paragraph{Latent queries and router head.}
The latent queries $\tilde{\mathbf{Q}}\in\mathbb{R}^{K\times D}$ are imported from
the ALAE encoder and kept frozen, so slot $k$ inherits the slot
semantics the autoencoder already learned.  They attend to the prompt memory
through two cross-attention blocks of the same design as the ALAE encoder's,
\[
  \mathbf{c} = \mathrm{LN}\bigl(\mathrm{CA}(\tilde{\mathbf{Q}},
      \mathbf{M}_{\mathrm{txt}})\bigr) \in \mathbb{R}^{K\times D},
\]
and a shared two-layer MLP of hidden width $512$ maps each context vector to the
per-prompt correction $\boldsymbol{\Delta}_k(y)\in\mathbb{R}^{L}$.  Its output
layer is zero-initialized.  We set $s_{\mathrm{s}} = s_{\mathrm{d}} = 4$, which
caps the logit margin at $8$.

\paragraph{Temperature schedule.}
During training the Gumbel-Softmax temperature is annealed linearly from
$\tau_{\mathrm{r}}{=}5$ to $1.5$ over the first $100$ epochs, so early training
explores soft mixtures over depth before the router commits.  Evaluation uses the
converged temperature rather than the annealed value: a detached inference pass
would otherwise read the schedule at epoch zero and route almost uniformly.

\paragraph{Mutual-information objective.}
Let $\mathbf{p}_k = \mathrm{softmax}(\mathbf{e}_k / \tau_{\mathrm{r}})$ be the
noise-free routing distribution of slot $k$, let $\mathcal{H}[\cdot]$ denote
entropy, and let $\bar{\,\cdot\,}$ denote the average over prompts in a batch.
Per slot we maximize
\[
  \mathcal{I}_{\mathrm{route}} =
  \mathcal{H}\bigl[\bar{\mathbf{p}}_k\bigr]
  - \overline{\mathcal{H}\bigl[\mathbf{p}_k\bigr]} ,
\]
weighted by $0.05$.  The first term pushes different prompts to route
differently, the second pushes each individual prompt to route decisively.  Both
are evaluated on $\mathbf{p}_k$ rather than on the sampled weights, since the
injected noise would inflate the conditional entropy artificially.  A plain
entropy bonus is not a substitute: it rewards soft mixtures for every prompt
instead of conditional switching.

\paragraph{Cost.}
The added cost over a standard rollout is the two-block cross-attention stack
over the $K$ latent queries plus the shared MLP.  The per-layer states $\mathbf{H}$
are already produced by the backbone's forward pass, and the routing logits
depend only on the prompt, so they are computed once per generation.

\paragraph{Per-verb routing.}
\begin{figure}[t]
    \centering
    \includegraphics[width=\columnwidth]{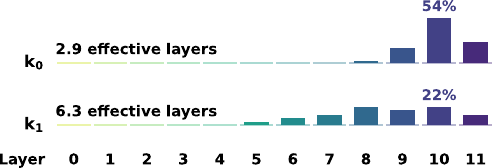}
    \caption{Layer-selection preference of the two slots on HumanML3D: $k_0$ is a narrow deep reader concentrated on layer~10, while $k_1$ is a wide scanner over layers~5--11.}
    \label{fig:router-slots}
\end{figure}
\begin{figure}[t]
    \centering
    \includegraphics[width=\columnwidth]{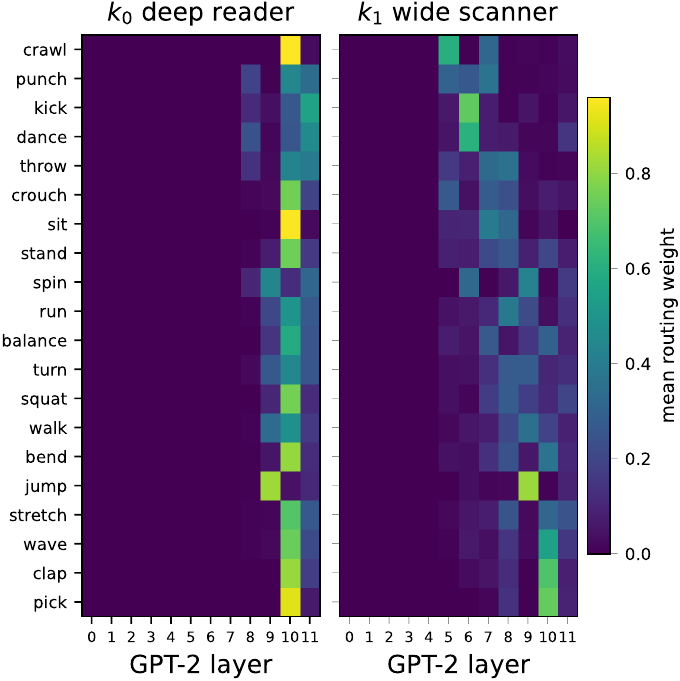}
    \caption{Per-verb layer preference on HumanML3D: $k_0$ (left) stays
    fixed on layer~10 for nearly every verb, whereas $k_1$ (right) shifts
    its preferred layer with the semantics of the action---shallow for
    ballistic/contact verbs (crawl, punch, kick), mid-depth for locomotion
    (run, walk, jump), and deep for object/hand actions (pick, clap, wave).}
    \label{fig:router-verbs}
\end{figure}
The scanner's depth also tracks verb semantics (Fig.~\ref{fig:router-verbs}),
reading shallow for ballistic/contact verbs and deep for object
manipulation---structure a collapsed static router cannot show.

\section{Factor Head Implementation Details}
\label{app:factor-head}

\paragraph{Parameterization.}
Each routed slot state $\tilde{\mathbf{h}}_k$ is layer-normalized and linearly
projected to the per-slot mean and diagonal scale, and the slot-pooled normalized
states produce the factor amplitudes.  The global basis
$\mathbf{U}\in\mathbb{R}^{KD\times r}$ is initialized with unit-norm columns.
The linear layer producing the log-amplitudes $\boldsymbol{\ell}(y)$ has its bias
initialized to $-2$, so $\mathbf{a}\approx e^{-2}$ and training starts near the
diagonal model, with the factor part earning its amplitude from the likelihood.

\paragraph{Closed-form likelihood.}
The negative log-likelihood of the low-rank-plus-diagonal family is computed in
closed form.  The Woodbury identity and the matrix determinant lemma reduce it to
diagonal operations plus one $r \times r$ Cholesky factorization per sample, so
the cost is linear in $KD$ rather than cubic, and no $KD \times KD$ covariance is
ever formed.

\paragraph{Decoupled anchor.}
Written out, the decoupled anchor $\mathcal{L}_{\mathrm{lat}}$ of the joint training
objective is
\[
\begin{array}{rcl}
  \mathcal{L}_{\mathrm{lat}} &=&
  -\log \mathcal{N}\bigl(\mathbf{z}^{\star};\,
      \boldsymbol{\mu},\,
      \mathrm{sg}[\mathrm{diag}(\boldsymbol{\sigma}^{2})]\bigr) \\[2pt]
  && -\,\log \mathcal{N}\bigl(\mathbf{z}^{\star};\,
      \mathrm{sg}[\boldsymbol{\mu}],\, \\[2pt]
  && \qquad\quad
      \mathbf{U}\mathrm{diag}(\mathbf{a})^{2}\mathbf{U}^{\top}
      + \mathrm{diag}(\boldsymbol{\sigma}^{2})\bigr),
\end{array}
\]
where $\mathrm{sg}[\cdot]$ denotes stop-gradient.  The first term trains only the
mean under a frozen diagonal precision, so the diagonal scale acts on the mean
update purely as a stop-gradient preconditioner.  The second fits the full
covariance to the detached residual.  The factor parameters
$(\mathbf{U},\mathbf{a})$ therefore never appear in any term that carries
gradient to $\boldsymbol{\mu}$, so the mean update is invariant to the covariance
fit.

\section{Joint Training Details}
\label{app:joint}

\paragraph{Loss weights.}
The decoupled anchor is weighted by $\lambda_{\mathrm{lat}}{=}2$ and the
motion-to-text term by $\lambda_{\mathrm{m2t}}{=}1$.

\paragraph{Motion understanding branch.}
The caption cross-entropy uses label smoothing $0.1$.  We perturb the input
latents at train time with zero-mean Gaussian noise of standard deviation
$0.05$, and we floor the caption loss at $2.8$, reflecting the gradient below
that value so the understanding direction cannot keep sharpening at the expense
of generation.  The unified-training ablation of App.~\ref{app:unified} shows
this floor in fact protects both directions.  Captions are decoded greedily
with at most $64$ new tokens.

\section{Sampling Protocol and Temperature Calibration}
\label{sec:sampling-protocol}

At evaluation time a motion latent is drawn in a single shot as
\[
  \begin{gathered}
  \bar{\mathbf{z}} =
  \boldsymbol{\mu}
  + \tau\,\bigl(
      \mathbf{U}(\mathbf{a}\odot\boldsymbol{\epsilon}_1)
      + \boldsymbol{\sigma}\odot\boldsymbol{\epsilon}_2
    \bigr),\\
  \boldsymbol{\epsilon}_1,\boldsymbol{\epsilon}_2 \sim \mathcal{N}(0,\mathbf{I}),
  \end{gathered}
\]
where the temperature $\tau$ scales all zero-mean perturbations around the mean
and $\tau{=}0$ recovers deterministic mean decoding (\emph{$\mu$-decode}).

Because the covariance is fitted by maximum likelihood on training residuals, it
is mildly over-dispersed on held-out text, so a post-hoc temperature slightly
below one is a legitimate calibration rather than a trick.  Our protocol is
strict about this: $\tau$ is selected \emph{only on the validation split} (a
single sweep over $\tau\in[0.5,1.0]$), and the test split is evaluated once at
the selected operating point, with $\tau{=}1.0$ and $\mu$-decode reported as
references on HumanML3D.

All quantitative claims use this \emph{honest-sampling protocol}: model selection is performed under sampled validation, every reported
metric, including diversity and MultiModality, comes from the same sampler at the
same operating point, and test numbers use the official 20-replication protocol
of each benchmark.

\section{Additional Experimental Results}
\label{app:results}

\subsection{$K$-Scaling on SnapMoGen}\label{app:kscaling-snap}

\begin{table}[t]
\centering
\small
\setlength{\tabcolsep}{2.6pt}
\begin{tabular}{@{}lcccccccc@{}}
\toprule
$K$ & $\tau^{*}$ & R@1$\uparrow$ & R@2$\uparrow$ & R@3$\uparrow$ & FID$\downarrow$ & CLIP$\uparrow$ & Div. & MM \\
\midrule
1  & 0.6 & 0.799 & 0.904 & 0.938 & 22.28 & 0.691 & 19.74 & 6.90 \\
4  & 0.6 & \textbf{0.815} & \textbf{0.916} & \textbf{0.947} & 21.05 & \textbf{0.698} & 19.79 & 6.09 \\
8  & 0.6 & 0.804 & 0.905 & 0.939 & 21.99 & 0.686 & 19.86 & 6.41 \\
16 & 0.6 & 0.801 & 0.904 & 0.939 & \textbf{20.76} & 0.686 & 19.86 & 7.28 \\
\bottomrule
\end{tabular}
\caption{Latent budget $K$ on the SnapMoGen test split, the companion of the
HumanML3D block reported in the main paper ($\tau^{*}$ =
validation-selected temperature, 20 replications, ground-truth Diversity
$19.5$--$19.8$).  Every retrieval metric and the CLIP score peak at
$K{=}4$, while
FID is best at $K{=}16$ over a spread of only $1.5$ points.}
\label{tab:kscaling-snap}
\end{table}

Table~\ref{tab:kscaling-snap} repeats the budget sweep of the main paper
on SnapMoGen.  The retrieval optimum transfers: a
small budget is best on both benchmarks, and leaving it costs precision.
The FID landscape does not, since the two benchmarks place their best and
worst FID budgets at opposite ends of the sweep.  The $K{=}16$ FID gain
here costs $1.4$ R@1 points, so it does not move the recommended operating
point away from $K{=}4$.

\subsection{Oracle Residual Diagnosis}\label{sec:oracle}

\begin{table}[t]
\centering
\small
\setlength{\tabcolsep}{4.5pt}
\resizebox{\columnwidth}{!}{%
\begin{tabular}{lcccccc}
\toprule
& $s{=}0$ & $s{=}0.25$ & $s{=}0.5$ & $s{=}0.75$ & $s{=}1$ & shuffle \\
\midrule
FID$\downarrow$ & 23.54 & 16.95 & 6.89 & 1.70 & 0.65 & 14.23 \\
R@1$\uparrow$ & 0.828 & 0.926 & 0.962 & 0.957 & 0.939 & 0.422 \\
\bottomrule
\end{tabular}}
\caption{Oracle residual decoding on SnapMoGen test: decode
$\mathbf{z} = \boldsymbol{\mu} + s\,(\mathbf{z}^{\star} -
\boldsymbol{\mu})$.  FID and R@1 improve \emph{together} along the true
residual direction; batch-shuffled residuals (right) collapse R@1, showing
the residual is text-conditional.  $s{=}1$ equals the stage-one
reconstruction ceiling.}
\label{tab:oracle}
\end{table}

To localize the remaining FID gap we decode
$\mathbf{z}=\boldsymbol{\mu}+s\,(\mathbf{z}^{\star}-\boldsymbol{\mu})$,
replacing the sampler with a scaled \emph{true} residual at zero training
cost (Table~\ref{tab:oracle}).  Three facts follow.  (i)~The representation
is not the bottleneck: with only half the true residual ($s{=}0.5$) FID
drops to $6.89$, far below the $20.8$--$22.3$ sampled FIDs of
Table~\ref{tab:kscaling-snap}---the representation retains
substantial oracle headroom.  The continuous slot representation, in other words, already contains
the motions the benchmark asks for.  (ii)~FID and R-Precision rise \emph{together} along the true
residual direction, suggesting that the FID--precision tradeoff observed
under isotropic or diagonal sampling families is an artifact of the sampling
family, not a property of the task.  (iii)~Batch-shuffling the residuals keeps FID moderate but collapses
R@1 to $0.42$: the missing variance is text-conditional and directionally
structured.  Together these pin the residual FID gap on the conditional
sampler---the single draw that single-shot generation allows---motivating the
structured factor head, and explaining why na\"ive noise injection cannot
close it.

\subsection{Latent Distribution Family}\label{app:ablation-head}

\begin{table}[t]
\centering
\small
\setlength{\tabcolsep}{5pt}
\begin{tabular}{lcc}
\toprule
Latent head (SnapMoGen $K{=}4$, val) & FID$\downarrow$ & R@1$\uparrow$ \\
\midrule
KL-pinned diagonal ($\sigma$ pinned at 1) & 25.8$^{\dagger}$ & -- \\
Diagonal NLL & 33.7 & -- \\
Cross-slot AR chain (sched.\ sampling) & 29.2 & -- \\
Factor head, joint ML (no decoupling) & 47.5$^{\ddagger}$ & 0.13$^{\ddagger}$ \\
Factor head, decoupled (ours) & \textbf{22.97} & \textbf{0.775} \\
\bottomrule
\end{tabular}
\caption{Distribution-family ablation on SnapMoGen. $^{\dagger}$Best value
across a $\tau\in[0,3]$ sweep---the KL-pinned model's temperature response
is nearly flat because training deletes text-conditional variance.
$^{\ddagger}$Joint ML collapses via variance-eats-gradient (shown at epoch
9, when the run was stopped).}
\label{tab:ablation-head}
\end{table}

\paragraph{Distribution family.}
Table~\ref{tab:ablation-head} traces the path to the factor head on
SnapMoGen.  The KL-regularized diagonal head cannot be rescued by test-time
temperature (a $\tau\in[0,3]$ sweep moves FID by ${<}3$ points): with the
variance pinned, training removes within-text variance from the mean
itself.  Replacing KL with a diagonal NLL calibrates $\sigma(y)$ but
plateaus hard at FID $33.7$---the diagonal independence assumption is the
binding constraint, consistent with the measured cross-slot correlations
($|\rho|\approx0.34$--$0.52$).  An autoregressive cross-slot chain restores
correlation but is brittle---and it surrenders the single-shot budget, replacing
the single draw with sequential sampling that peaks at $29.2$ (epoch 49) and
then overfits its own conditional mechanism.  The single-shot factor head
reaches $22.97$, with \emph{rising} R@1 indicating sampling in the correct
residual directions (App.~\ref{sec:oracle}).

\paragraph{Decoupled optimization is necessary.}
Under naively joint maximum likelihood the factor amplitudes absorb the
mean error within a few epochs ($\bar{a}$: $0.15\rightarrow4.2$) while
retrieval collapses to near-random---the variance-eats-gradient failure.
The decoupled anchor removes the failure without clamps: the mean gradient
is bit-identical regardless of the covariance parameters, and the amplitude
trajectory becomes rise-then-fall, tracking the shrinking residual.

\subsection{Sampling versus $\mu$-decode}\label{app:sampling-ablation}

\begin{table}[t]
\centering
\small
\setlength{\tabcolsep}{3pt}
\resizebox{\columnwidth}{!}{%
\begin{tabular}{lcccccc}
\toprule
Decoding & R@1$\uparrow$ & R@2$\uparrow$ & R@3$\uparrow$ & FID$\downarrow$ & MM-Dist$\downarrow$ & MM \\
\midrule
$\mu$-decode & \textbf{0.585} & \textbf{0.778} & \textbf{0.863} & 0.131 & \textbf{2.611} & -- \\
$\tau{=}1.0$ & 0.555 & 0.751 & 0.844 & 0.151 & 2.735 & 1.35 \\
$\tau^{*}{=}0.6$ & 0.579 & 0.772 & 0.859 & \textbf{0.087} & 2.629 & 0.83 \\
\bottomrule
\end{tabular}}
\caption{Sampling-protocol ablation on the HumanML3D test set: deterministic
$\mu$-decode versus uncalibrated ($\tau{=}1.0$) and validation-calibrated
($\tau^{*}{=}0.6$) sampling. Calibrated sampling trades ${\approx}0.6$ R@1
points for a $34\%$ lower FID than $\mu$-decode.}
\label{tab:hml-sampling}
\end{table}

Table~\ref{tab:hml-sampling} isolates the sampling protocol on HumanML3D.
The single calibrated draw improves FID by $34\%$ over deterministic
$\mu$-decode ($0.131\rightarrow0.087$) at a small retrieval cost
($0.585\rightarrow0.579$ R@1), and the temperature exposes an explicit
diversity--fidelity dial (MM $0.83\rightarrow1.35$ from $\tau{=}0.6$ to
$1.0$).  On SnapMoGen the FID
dividend is smaller and likewise costs ${\approx}0.01$ R@1---consistent with the
smaller dataset leaving a more over-dispersed ML covariance.  In both cases
$\tau{=}1.0$ is never optimal, confirming that validation-side calibration
is a necessary component of an honest-sampling protocol rather than an
optional refinement.

Across latent budgets on both benchmarks, the
HumanML3D $\mu$-decode$\rightarrow$best-$\tau$ FID dividend is
$0.026$--$0.030$ for $K{\leq}4$ and $0.064$--$0.078$ for $K{\geq}8$.  On
SnapMoGen the validation-selected temperature is $0.6$ at every budget
under the protocol's R@1 selection rule, and best-temperature sampling
beats $\mu$-decode by $2.0$--$5.9$ validation-FID points
($26.4\rightarrow23.3$ at $K{=}4$).

\subsection{Motion Understanding on SnapMoGen}\label{app:snap-m2t}

\begin{table}[t]
\centering
\small
\setlength{\tabcolsep}{5pt}
\begin{tabular}{lcc}
\toprule
SnapMoGen test & \sysname~(Ours) & Real captions \\
\midrule
R@1$\uparrow$          & 0.600 & 0.945 \\
R@2$\uparrow$          & 0.750 & 0.980 \\
R@3$\uparrow$          & 0.819 & 0.988 \\
CLIP score$\uparrow$   & 0.580 & 0.840 \\
\midrule
BLEU@1$\uparrow$       & 64.655 & -- \\
BLEU@4$\uparrow$       & 23.492 & -- \\
ROUGE$\uparrow$        & 37.675 & -- \\
CIDEr$\uparrow$        & 9.392  & -- \\
BERTScore$\uparrow$    & 26.354 & -- \\
\bottomrule
\end{tabular}
\caption{Motion understanding on the SnapMoGen test set, from the same
flagship checkpoint and the same operating point as the SnapMoGen
generation results of the main paper, under the official 20-replication
protocol.  The
right column is the real-caption reference measured in the same run, which
upper-bounds the retrieval columns.  All 95\% confidence intervals are below
$\pm0.002$ on the retrieval columns and below $\pm0.05$ on the captioning
columns.  We are not aware of published captioning results on SnapMoGen, so
no baseline row is available.}
\label{tab:snap-m2t}
\end{table}

Table~\ref{tab:snap-m2t} reports the understanding direction on SnapMoGen.
The captioning branch is trained jointly with generation and reads the same
frozen latent slots, so these numbers come from the SnapMoGen generation
checkpoint of the main paper at no additional training cost.  One protocol
difference from HumanML3D is worth stating: the SnapMoGen checkpoint is
selected by FID alone (App.~\ref{app:setup}), so unlike the HumanML3D
flagship it receives no understanding-side model selection.

\paragraph{Retrieval.}
Generated captions retrieve their own motion at R@1 $0.600$, well below the
$0.945$ of the human-written captions.  This is the opposite of the
HumanML3D picture, where the same branch retrieves \emph{above} the
real-caption reference at every rank, and it
mirrors the generation side: SnapMoGen is the harder benchmark in both
directions for a single shared checkpoint.

\paragraph{Why the captioning scores are not comparable across benchmarks.}
A SnapMoGen clip carries six reference captions averaging $49$ words with
body-part-level detail, against three references averaging $13$ words on
HumanML3D.  Our SnapMoGen captions match that length (median $49$ words), so
the gap is not a brevity artifact.  Under those long, highly specific
references $n$-gram precision stays high, in fact above the HumanML3D values
reported in the main paper (BLEU@1 $64.7$ vs.\ $60.8$, BLEU@4 $23.5$ vs.\
$22.0$), while the consensus-weighted and embedding-based scores fall sharply
(CIDEr $9.4$ vs.\ $50.4$, BERTScore $26.4$ vs.\ $43.4$).  The two families
disagree because they weight different things: the model produces fluent,
correctly scaled descriptions of the coarse motion, and it misses the rare,
high-information detail that carries the TF-IDF weight in CIDEr.  Captioning
scores should therefore be read within a benchmark, never across the two.

\subsection{Motion Generation Case Study}\label{app:t2m-cases}

\begin{table*}[t]
\centering
\footnotesize
\setlength{\tabcolsep}{4pt}
\setlength{\abovecaptionskip}{4pt}
\renewcommand{\arraystretch}{1.0}
\begin{tabular}{@{}l>{\centering\arraybackslash}m{0.271\textwidth}>{\centering\arraybackslash}m{0.271\textwidth}>{\centering\arraybackslash}m{0.271\textwidth}@{}}
\toprule
Prompt
  & \csq{person \gcue{stands still} and aggressively \gcue{points straight forward}}
  & \csq{a person who is standing on a balance beam \gcue{takes three steps forward} and then \gcue{steps down off the beam}.}
  & \csq{a person starts a \gcue{jogging on the place}} \\
\midrule
MotionGPT3        & \csgen{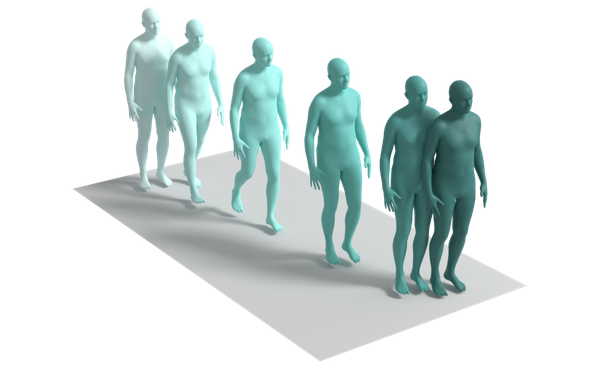}{2.55} & \csgen{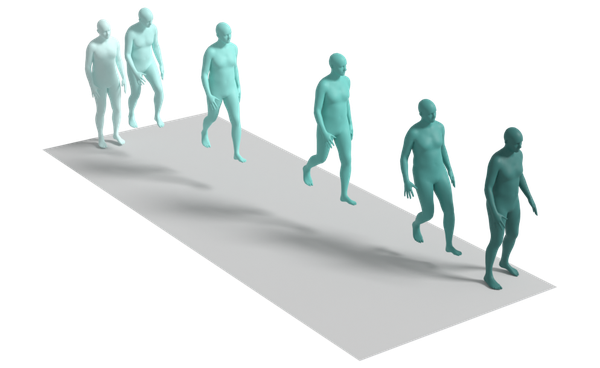}{5.10} & \csgen{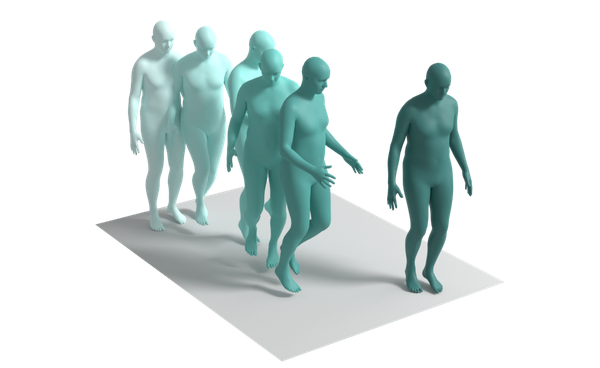}{1.59} \\
\midrule
MoMask++          & \csgen{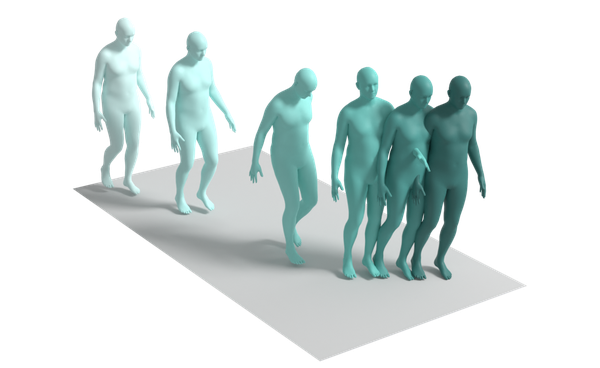}{2.30} & \csgen{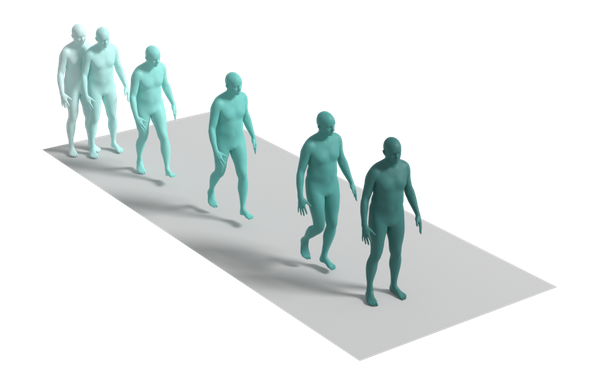}{4.68} & \csgen{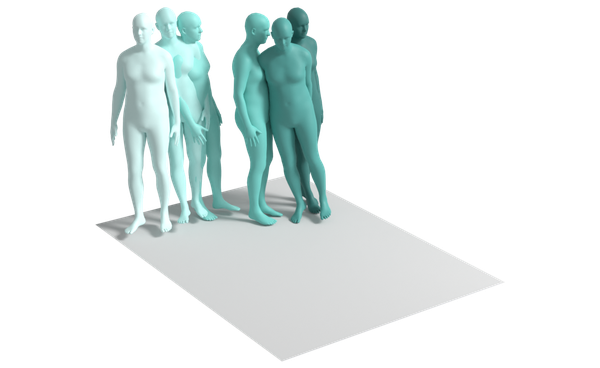}{0.54} \\
\midrule
\sysname{} (Ours) & \csgen{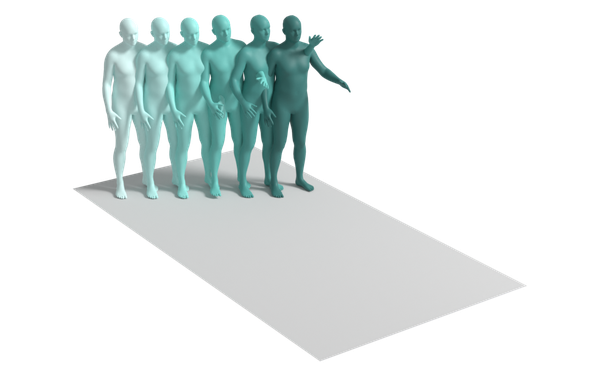}{0.04}       & \csgen{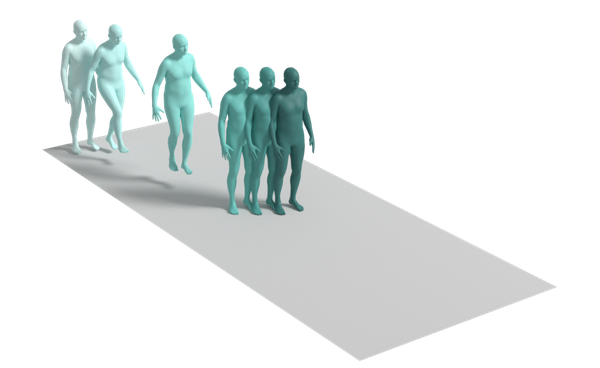}{2.32}       & \csgen{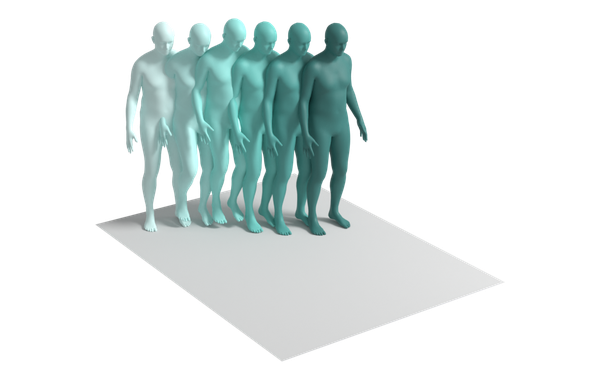}{0.15}       \\
\midrule
Real              & \csgen{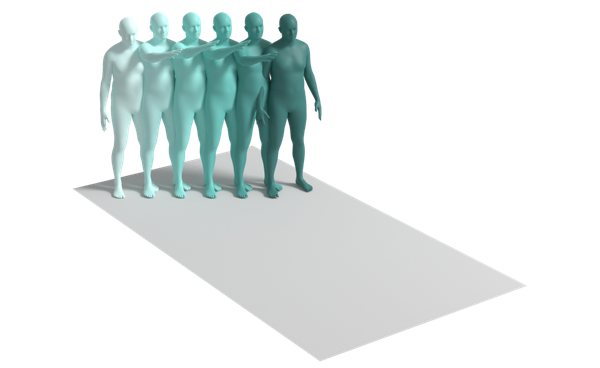}{0.04}         & \csgen{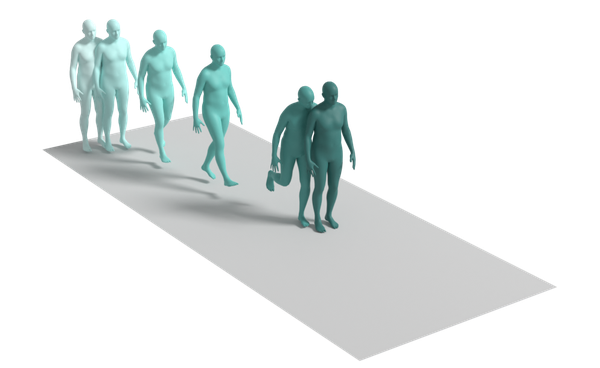}{2.94}         & \csgen{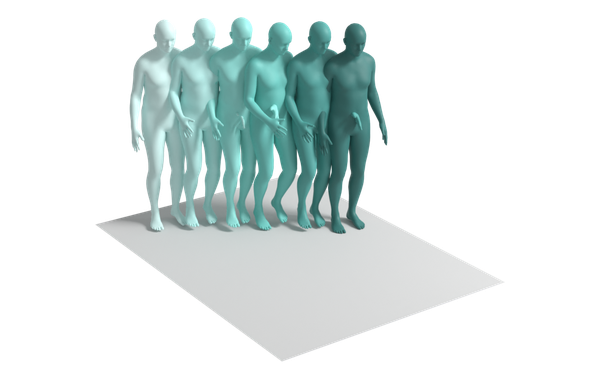}{0.10}         \\
\bottomrule
\end{tabular}
\caption{Motion generation case study on HumanML3D test.  All three systems
receive the \emph{same} prompt and the \emph{same} target length as the
reference motion; poses are rendered light to dark over time, and the four
rows of a column share one camera and floor tile, so sizes may be compared
vertically.  The number under each render is the root horizontal displacement
$d$ in metres.  Each prompt constrains how far the body should travel
(\gcue{teal}), and both baselines overshoot that constraint on every clip
shown: on \csq{stands still} they cover $2.30$ and $2.55$\,m where the
reference covers $0.04$\,m, and on \csq{jogging on the place} MotionGPT3
drifts $1.59$\,m against $0.10$\,m.  Averaged over the five clips, the
deviation $|d-d_{\text{real}}|$ is $0.19$\,m for \sysname{}, $1.32$\,m for
MoMask++ and $2.09$\,m for MotionGPT3.  The clips are drawn from the $128$
test motions that are non-inferior to \emph{both} baselines on every physical
axis we measure, a pool fixed before any figure was chosen; the per-clip
displacement gap is therefore illustrative of that regime and is not an
aggregate claim.}
\label{tab:t2m-cases}
\end{table*}

\begin{table*}[t]\ContinuedFloat
\centering
\footnotesize
\setlength{\tabcolsep}{4pt}
\setlength{\abovecaptionskip}{4pt}
\renewcommand{\arraystretch}{1.0}
\begin{tabular}{@{}l>{\centering\arraybackslash}m{0.271\textwidth}>{\centering\arraybackslash}m{0.271\textwidth}>{\centering\arraybackslash}m{0.271\textwidth}@{}}
\toprule
Prompt
  & \csq{a person \gcue{walks forward} and \gcue{turns to the right}.}
  & \csq{a person \gcue{walks up to a table} and starts \gcue{washing it}.}
  & \\
\midrule
MotionGPT3        & \csgen{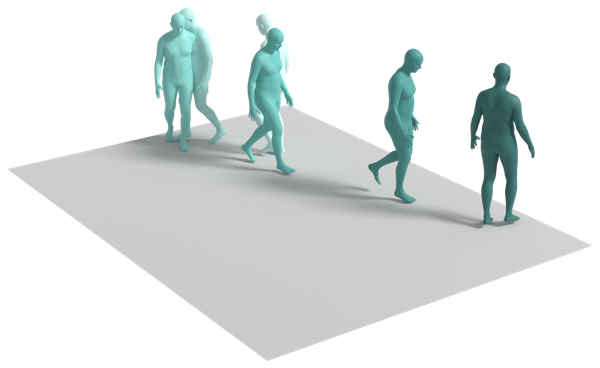}{3.82} & \csgen{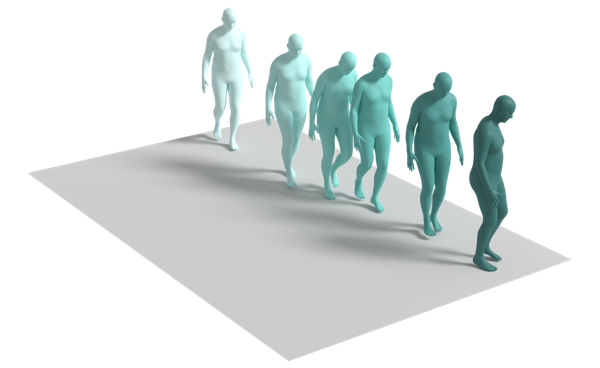}{3.56} & \\
\midrule
MoMask++          & \csgen{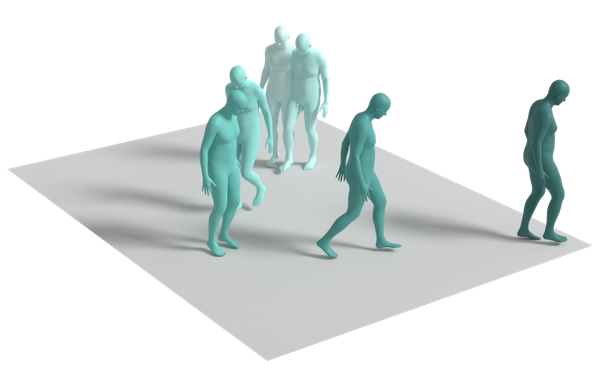}{3.61} & \csgen{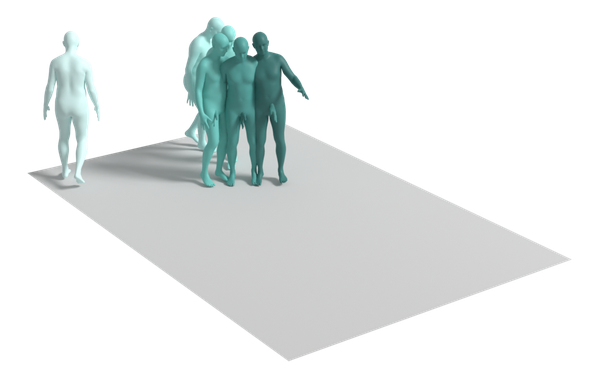}{1.67} & \\
\midrule
\sysname{} (Ours) & \csgen{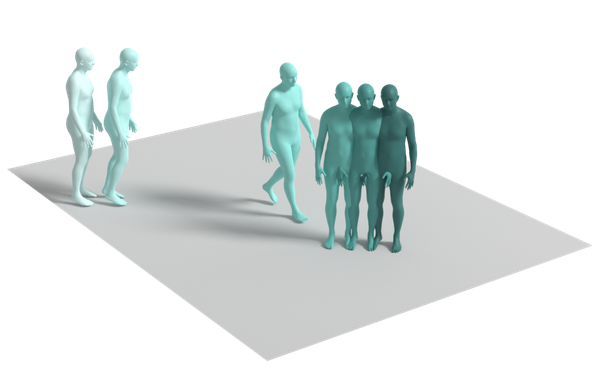}{2.32}       & \csgen{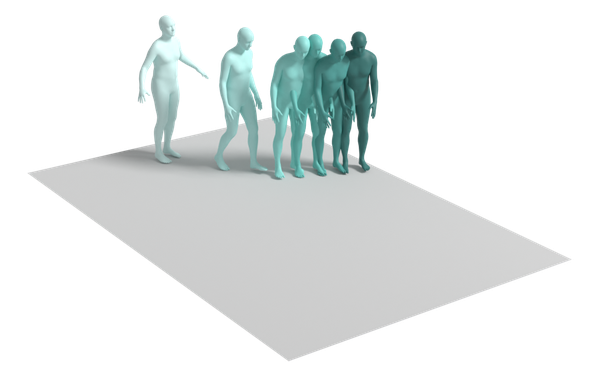}{1.00}       & \\
\midrule
Real              & \csgen{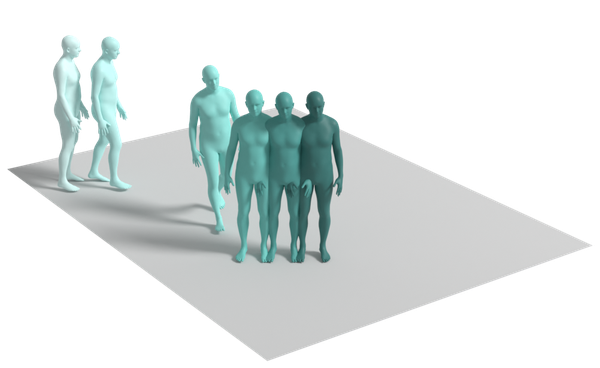}{2.34}         & \csgen{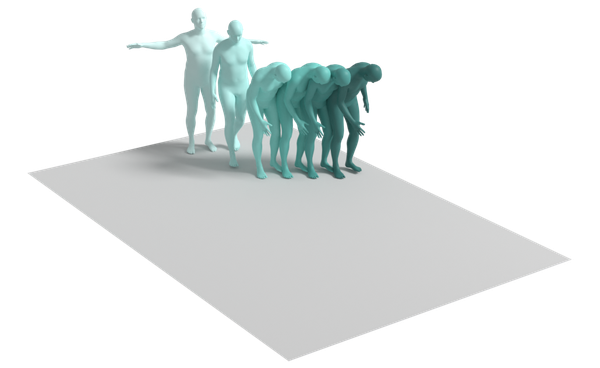}{0.76}         & \\
\bottomrule
\end{tabular}
\caption{Motion generation case study on HumanML3D test (continued).  On
\csq{walks forward and turns to the right} MotionGPT3 not only travels
$3.82$\,m against the reference $2.34$\,m but also turns through $277^\circ$
where the reference turns $91^\circ$ and \sysname{} through $104^\circ$, so
its final pose faces away from the camera.}
\label{tab:t2m-cases-b}
\end{table*}

Table~\ref{tab:t2m-cases} makes the aggregate HumanML3D generation numbers
of the main paper concrete on individual motions, and shows that the
two baselines fail in different directions.  MotionGPT3 converts stationary
prompts into locomotion: it travels further than the reference on all five
clips, most starkly on \csq{stands still}, where the render shows a full
walking sequence.  MoMask++ follows trajectories more faithfully---measured
over the whole test set its trajectory error is statistically
indistinguishable from ours---but its poses carry visible high-frequency
noise, consistent with a jitter of $1.64\times$ the real level; MotionGPT3
sits at $0.61\times$ and is correspondingly over-smoothed.  \sysname{} is the
only one of the three that stays close to the reference on both quantities at
once.  We emphasise that lower jitter is not by itself better: MotionGPT3
attains the lowest absolute jitter of the three while being the furthest from
real motion on that axis, so the quantity to read is the distance to the
reference rather than the raw value.

\subsection{Motion Understanding Case Study}\label{app:m2t-cases}

\begin{table*}[t]
\centering
\footnotesize
\setlength{\tabcolsep}{4pt}
\setlength{\abovecaptionskip}{4pt}
\renewcommand{\arraystretch}{1.0}
\begin{tabular}{@{}l>{\centering\arraybackslash}p{0.271\textwidth}>{\centering\arraybackslash}p{0.271\textwidth}>{\centering\arraybackslash}p{0.271\textwidth}@{}}
\toprule
Input Motions & \csmot{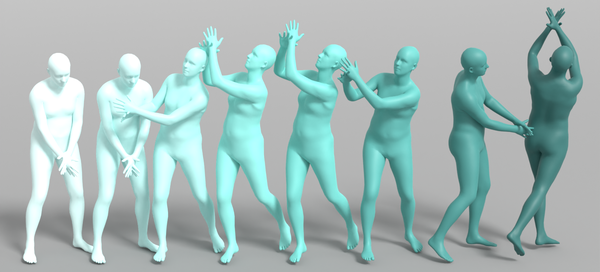} & \csmot{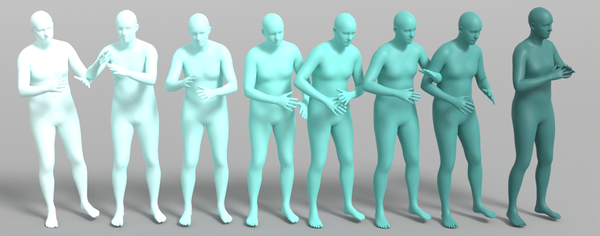} & \csmot{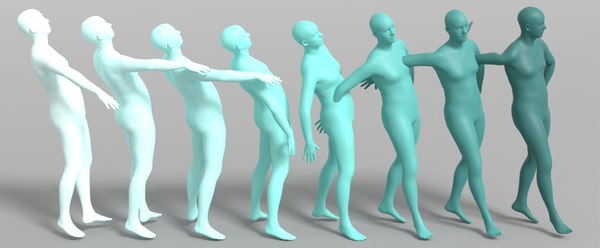} \\
\midrule
Real & \csq{the person \gmatch{swings a golf club}} & \csq{a person \gmatch{sweeping the floor}} & \csq{a person underhand \gmatch{throws something with their right hand}.} \\
\midrule
MotionGPT3 & \csq{a man holds both hands together near his groin and then simultaneously lifts them up in front of him.} & \csq{a person uses a hammer to swing an object, while holding it in its hands.} & \csq{a person reaches right hand up and swings it in front of them} \\
\midrule
\sysname{} (Ours) & \csq{a person \gmatch{swings a golf club}.} & \csq{a person is \gmatch{sweeping the floor}.} & \csq{a person \gmatch{throws something with their right hand}.} \\
\midrule[\heavyrulewidth]
Input Motions & \csmot{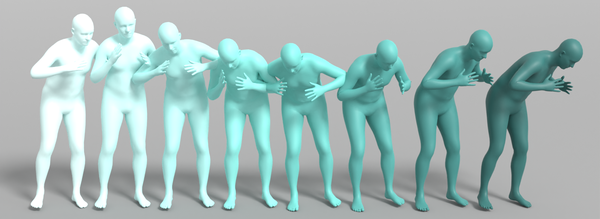} & \csmot{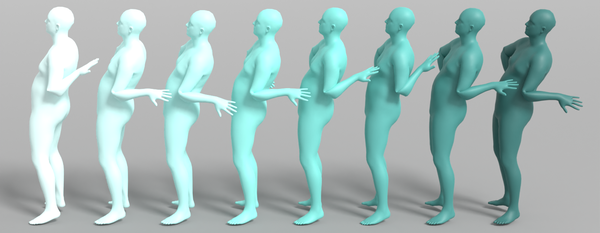} & \csmot{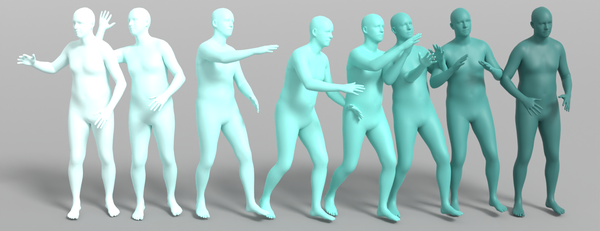} \\
\midrule
Real & \csq{the person \gmatch{is acting like a} human \gmatch{chicken}.} & \csq{the person \gmatch{is lifting a dumbbell with the left hand}.} & \csq{a person \gmatch{throws something with their right hand}.} \\
\midrule
MotionGPT3 & \csq{a person bends over and rubs their hands together.} & \csq{a person bends their arm at the elbow repeatedly} & \csq{a person throws something, then catches it with both hands.} \\
\midrule
\sysname{} (Ours) & \csq{a person \gmatch{is acting like a chicken}.} & \csq{a person \gmatch{is lifting a dumbbell with their left hand}.} & \csq{a person \gmatch{throws something with their right hand}.} \\
\midrule[\heavyrulewidth]
Input Motions & \csmot{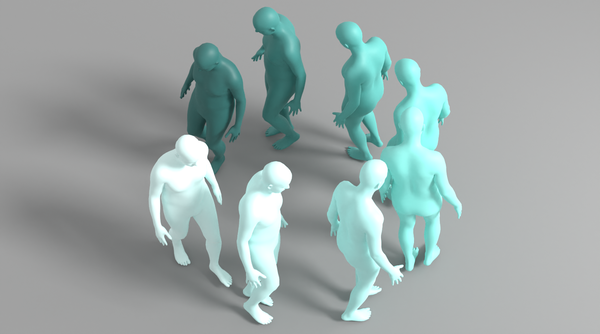} & \csmot{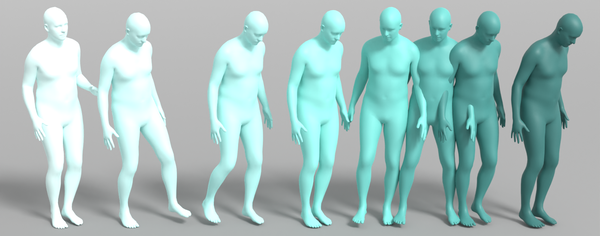} & \csmot{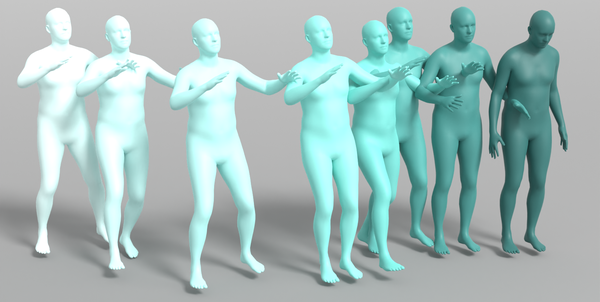} \\
\midrule
Real & \csq{person \gmatch{is walking} normally \gmatch{in a circle}} & \csq{a person \gmatch{side steps to the left and then to the right}.} & \csq{the person \gmatch{is dancing the waltz}.} \\
\midrule
MotionGPT3 & \csq{a man walks forward in an oval pattern} & \csq{a person slightly moved forward and return} & \csq{a person is dancing with a box step.} \\
\midrule
\sysname{} (Ours) & \csq{a person \gmatch{walks in a circle}} & \csq{a person \gmatch{steps to the left, then steps back to the right}.} & \csq{a person \gmatch{is dancing the waltz}.} \\
\bottomrule
\end{tabular}
\caption{Motion understanding case study on HumanML3D test.  Each block
shows the input motion (rendered light to dark over time), a reference
caption, and both models' captions; phrases where \sysname{} agrees with
the reference are marked in \gmatch{teal}.  The MotionGPT3 captions admit
no such alignment: the action, the object, or the trajectory is replaced
or lost.}
\label{tab:m2t-cases}
\end{table*}

Table~\ref{tab:m2t-cases} illustrates the understanding gap behind the
aggregate HumanML3D metrics of the main paper.  The failures on the
MotionGPT3 row are not paraphrase noise: the model replaces the action
with a different one, drops the object that defines it, or loses the
trajectory, while low-order $n$-grams (\csq{a person \ldots}) still match.
This is the microscopic account of why the margin between the two systems
widens with metric order, from $+1.7$ BLEU@1 to $+2.6$ BLEU@4 to $+21.7$
CIDEr in the aggregate metrics.

\clearpage

\subsection{Language Backbone Ablation}\label{app:backbone}

\paragraph{Setup.}
This ablation varies the stage-two language backbone and nothing else.  All
five runs use the $K{=}2$ HumanML3D flagship recipe of the main paper:
the same frozen stage-one ALAE checkpoint, factor
rank $r{=}64$, the depth router, the joint captioning
branch, 500 epochs, effective batch 256, and backbone learning rate
$10^{-4}$.  Each configuration is derived from the flagship configuration
file by changing only the backbone path (and, for the non-GPT-2 runs, the
precision and the adapter block), so the five columns are key-for-key
identical elsewhere.  The router reads the backbone depth from the model
configuration, which gives $L{=}12$ layers for GPT-2, $L{=}28$ for both
Qwen3 models, and $L{=}24$ for SmolLM2.

The backbones are GPT-2 124M \cite{gpt2}, Qwen3-0.6B and Qwen3-1.7B
\cite{qwen3}, and SmolLM2-1.7B \cite{smollm2}.  The two Qwen3 sizes isolate
model scale within a single family and tokenizer, while SmolLM2 matches
Qwen3-1.7B in parameter count but comes from an independent pretraining
pipeline.  The billion-scale backbones are adapted with LoRA \cite{lora}
($r{=}16$, $\alpha{=}32$, dropout $0.05$) on the seven projection matrices
of every block; the motion-token embedding rows added to the vocabulary stay
fully trainable and are excluded from weight decay, as in the flagship.  Two
departures from the flagship are forced by memory: these runs use bf16 mixed
precision, because a 500-epoch fp32 run at this scale would take several
days, and a micro-batch of 128 with two gradient-accumulation steps, because
a micro-batch of 256 exceeds the memory of a single B200 at 1.7B parameters.
Mixed precision is itself a confound, so we add a bf16 GPT-2 twin as a
precision-matched control, and every backbone claim below is read against
that twin rather than against the fp32 flagship.  These runs use
Transformers 4.51, the minimum version that supports Qwen3.  Model
selection, temperature calibration and the test protocol follow
App.~\ref{app:setup}: an eight-point validation temperature sweep,
$\tau^{*}$ at maximal validation R@1, and the official 20-replication test
protocol at $\tau^{*}$.  Stage-two wall-clock on one B200 was $5.3$\,h
(GPT-2 bf16), $18.1$\,h (Qwen3-0.6B), $18.8$\,h (SmolLM2-1.7B) and
$26.1$\,h (Qwen3-1.7B).

\begin{table}[t]
\centering
\footnotesize
\newcommand{\bkci}[2]{#1{\scriptsize$\pm#2$}}
\setlength{\tabcolsep}{2.2pt}
\resizebox{\columnwidth}{!}{%
\begin{tabular}{@{}lccccc@{}}
\toprule
Backbone & FID$\downarrow$ & R@1$\uparrow$ & B@4$\uparrow$ & CIDEr$\uparrow$ & Joint$\uparrow$ \\
\midrule
GPT-2 124M$^{\dagger}$ & \bkci{0.087}{.004} & \bkci{0.579}{.002} & 22.0 & \textbf{50.4} & \textbf{66.8} \\
GPT-2 124M$^{\ddagger}$ & \bkci{0.125}{.005} & \bkci{0.575}{.003} & \textbf{22.5} & 50.1 & 66.3 \\
Qwen3-0.6B & \bkci{\textbf{0.080}}{.002} & \bkci{0.579}{.003} & 20.0 & 46.8 & 65.3 \\
Qwen3-1.7B & \bkci{0.091}{.003} & \bkci{\textbf{0.585}}{.002} & 21.3 & 48.3 & 66.4 \\
SmolLM2-1.7B & \bkci{0.147}{.004} & \bkci{0.564}{.003} & 20.3 & 48.0 & 64.2 \\
\bottomrule
\end{tabular}}
\caption{Language backbone ablation on the HumanML3D test set, all rows at
the best-joint checkpoint and its calibrated $\tau^{*}$.
$^{\dagger}$fp32, the flagship backbone;
$^{\ddagger}$bf16, the precision-matched control.  Every row except the
flagship uses bf16 mixed precision, so the three billion-scale backbones
should be read against that control rather than against the flagship.
FID and R@1 are means with 95\% confidence intervals over
the official 20 replications; captioning is deterministic greedy decoding
evaluated once, so B@4, CIDEr and the joint generation--understanding score
carry no interval, and B@4 and CIDEr follow the $\times100$ scale of the
main paper.}
\label{tab:backbone}
\end{table}

\paragraph{The backbone matters, but not through its size.}
At equal parameter count the two 1.7B backbones are far apart:
SmolLM2 is $61\%$ worse in FID than Qwen3 ($0.147$ vs.\ $0.091$, disjoint
intervals), $0.021$ lower in R@1, and $2.1$ points lower on the joint score.
Ordering Table~\ref{tab:backbone} by parameter count instead gives $0.125$
(124M), $0.080$ (0.6B), $0.091$ (1.7B) and $0.147$ (1.7B), which is not
monotone.  What a backbone contributes is therefore a property of the
pretrained model as a whole rather than of its capacity.  The two 1.7B
backbones also differ in tokenizer, vocabulary size ($49$k vs.\ $152$k) and
depth (24 vs.\ 28 layers), so we read the gap as a model-level effect and do
not attribute it to pretraining data alone.

\paragraph{Checkpoint selection moves FID as much as the backbone does.}
Selecting by best validation FID instead of the flagship's joint criterion
leaves Qwen3-1.7B and SmolLM2 unchanged, since for both the two criteria
pick the same epoch, but it moves Qwen3-0.6B from $0.080$ (epoch 269) to
$0.117$ (epoch 329), a spread of $0.036$ inside a single run.  That spread
is as large as the between-backbone differences at the top of the table, and
every configuration here is a single seed, so we do not rank backbones by
FID gaps of a few thousandths and we do not promote the nominally best FID
of the table into a headline number.  The joint score is the more stable
readout, because it constrains both directions at once and is insensitive to
noise in either one alone.

\paragraph{Captioning does not improve with a larger backbone.}
No swap helps the understanding direction: CIDEr is highest for the two
GPT-2 runs ($50.5$ and $50.1$) and lower for every billion-scale backbone
($46.8$ to $48.3$), and BERTScore F1 is likewise highest for the fp32
GPT-2 run ($43.7$), with every other backbone between $40.2$ and $43.2$.  The captioning
branch reads frozen ALAE latents through a projector, so its bottleneck is
the motion representation rather than language-model capacity, which is
consistent with the latent-side ablations of App.~\ref{app:ablation-head}.

\paragraph{Why the flagship keeps GPT-2.}
Qwen3-1.7B is the strongest alternative on the generation side, with the
best R@1 in the table and an FID $27\%$ below its precision-matched twin,
but that advantage does not carry over to the joint objective the flagship
is selected on, where GPT-2 remains highest.  Combined with the captioning
ordering above, the $14\times$ parameter cost, and a stage-two run that is
five times longer, we keep GPT-2 as the backbone of the reported system and
report this study as evidence that the pipeline is not bottlenecked by
language-model capacity.

\subsection{Unified-Training Ablation}\label{app:unified}

\paragraph{Setup.}
This ablation varies the training objective and nothing else.  All runs use
the $K{=}2$ HumanML3D flagship recipe (App.~\ref{app:backbone}, first row)
and are derived from the flagship configuration file by changing only the
keys named below, so the rows of the unified-training table of the main
paper are key-for-key identical elsewhere.  \emph{Generation only} disables the captioning task,
so the motion-to-text branch is never built and the backbone is trained by
the generation objective alone.  \emph{Understanding only} zeroes every
generation-side loss term, so the backbone is trained by caption
cross-entropy alone; the caption-loss floor of App.~\ref{app:joint} is
removed as well, since there is no generation left to protect.
\emph{Joint, no floor} keeps both objectives at
$\lambda_{\mathrm{m2t}}{=}1$ and removes only the floor.  The last two rows
move $\lambda_{\mathrm{m2t}}$ to $0.25$ and $4$ with the floor kept.  Model
selection uses the only criterion available to each regime: best validation
FID for generation only, the understanding side of the joint score for
understanding only, and the flagship's joint criterion for the joint rows.
Temperature calibration and the test protocol follow App.~\ref{app:setup};
every swept row calibrates to $\tau^{*}{=}0.6$, and the understanding-only
row needs no calibration because captions are decoded greedily.  Each
configuration is a single seed trained for the full 500 epochs on one B200
($3.7$\,h for generation only, $5.2$ to $5.6$\,h otherwise), so as in
App.~\ref{app:backbone} we do not read FID gaps of a few thousandths.  In
that table, FID and R@1 are means with 95\% confidence
intervals over the official 20 replications.  CIDEr and BERTScore F1
carry no interval: every row uses deterministic greedy decoding
evaluated once, except the flagship row, which quotes the
20-replication harvest of the main understanding table.  The Params
column counts all trained parameters of each checkpoint, excluding
evaluation-only machinery.

\paragraph{The generation gain is not a selection artifact.}
The generation-only row is selected by best
validation FID, the flagship by the joint criterion of
App.~\ref{app:setup}.  Selecting the flagship run by validation FID as
well, the same criterion the generation-only row uses, still yields
$0.090$, so the FID gap read in the main paper ($0.107$ vs.\
$0.087$) survives criterion-matched selection.

\paragraph{A dedicated captioner is barely better and does not survive the
schedule.}
The understanding-only run reaches CIDEr $52.8$ and BERTScore F1 $44.8$,
but its validation score improves for the last time at epoch $29$ and never
again over the remaining $470$ epochs: with no floor and no competing
objective the caption loss keeps falling while held-out quality stalls,
the signature of memorization.  The flagship stays within $5\%$ of it in
CIDEr ($50.4$) while being selected at epoch $239$, and at
$\lambda_{\mathrm{m2t}}{=}0.25$ the joint model reaches $53.5$, nominally
above the dedicated model; given single-seed selection noise we read this
as parity.  Sharing the backbone thus costs the understanding direction
little to nothing, and removes the need for the early stopping the
dedicated model depends on.

\paragraph{The floor protects both directions.}
The floor was introduced to keep the understanding branch from sharpening
at the expense of generation (App.~\ref{app:joint}), but the naive-joint
row shows its effect is not a one-sided trade: removing it degrades
generation (FID $0.100$, disjoint from $0.087$) and degrades understanding
even more (CIDEr $40.7$, a $19\%$ drop; BERTScore $39.2$).  An
unconstrained caption term drags the shared backbone toward caption
memorization and thereby hurts its own held-out captioning as well.  The
floor is best understood as a regularizer for both directions rather than
as protection for one.

\paragraph{The operating point is insensitive to
$\lambda_{\mathrm{m2t}}$ over a decade.}
$\lambda_{\mathrm{m2t}}{=}0.25$ and $1$ are statistically indistinguishable
on generation ($0.089$ vs.\ $0.087$, overlapping intervals) and comparable
on understanding, while $\lambda_{\mathrm{m2t}}{=}4$ degrades generation
($0.111$, disjoint from $0.087$) without helping understanding ($49.7$).
The useful range is therefore wide below the default, and the upper bound
lies between $1$ and $4$.

\subsection{Computational Requirements}\label{app:compute}

\paragraph{Protocol.}
The inference-cost table of the main paper reports
what it costs to generate one motion with
each of the three systems whose HumanML3D generations are compared in
App.~\ref{app:t2m-cases}.  All three run on one NVIDIA L4 GPU, from their
public checkpoints, over the same pinned list of captions and target
lengths, and each at the operating point of its own reported results:
MoMask++ at conditioning scale $5$ with $18$ masked-decoding steps,
MotionGPT3 at guidance scale $7.5$, and \sysname{} at $\tau^{*}{=}0.6$
without classifier-free guidance.  Each row is therefore the cost of
reaching that row's published quality, not a cost at some artificially
equalized setting.  Timing covers text to motion features on the GPU and
excludes forward kinematics, denormalization and file I/O, which all three
share.  Latency is the median over $64$ single-sample generations after
warm-up, and throughput uses batches of $32$.  Parameter counts include the
frozen text encoder each system loads at inference, a T5-v1.1-base encoder
($109.6$M) for MoMask++ and the language backbone itself for the other two.
Target lengths are capped at $196$ frames for every system because MoMask++
has a fixed positional grid at that length.

\paragraph{Discussion.}
The compute column and the latency column of that table do not
rank the two baselines the same way.  MotionGPT3 needs
$8.6\times$ our FLOPs but $14.3\times$ our latency, so its serial decoding is
latency-bound rather than compute-bound, while MoMask++ needs $9.5\times$ our
FLOPs and only $5.8\times$ our latency because its masked steps are large
dense products that keep the device busy.  At batch $32$ the per-motion
factors are $5.9$ and $6.1$.

\paragraph{Scope of the claim.}
The saving is in inference steps, not in model size.  At $222.3$M parameters
\sysname{} is larger than MoMask++ ($150.2$M) and smaller than MotionGPT3
($299.7$M), and its peak memory at batch $32$ ($1.66$\,GB) likewise sits
between the two ($0.73$ and $3.50$\,GB), because decoding a clip in one pass
materializes the whole feature tensor.  The measurement is also specific to
HumanML3D: MoMask++ spends the same $105.5$ GFLOPs on every sample since its
token grid and text padding are fixed, while our cost varies with caption
length ($9.9$ to $12.1$ GFLOPs across the sampled clips), so the ratios would
move on a benchmark with longer motions.  Training cost is likewise outside
this table.  All three rows evaluate public checkpoints and only ours was
trained by us, so baseline training budgets are stated as published
configurations rather than measured GPU-hours: MoMask++ trains a residual VQ
for $1000$ epochs at batch $256$ and a mask transformer for $500$ epochs at
batch $64$, and MotionGPT3 uses a three-stage schedule.  Our own budget is
given in App.~\ref{app:setup}.

\subsection{What Did Not Help}\label{app:negative}

A set-level mixture extension of the factor head ($M{=}8$ tied-covariance
components) trained stably, but its mixture component was a consistent
\emph{net negative} on test FID at every temperature we evaluated (e.g.\
$22.07$ vs.\ $21.18$ at the same $\tau$ with the mixture disabled).  The
unimodal factor sampler remains our best configuration; realizing the oracle
headroom in Table~\ref{tab:oracle} remains open.

\end{document}